\documentclass[10pt,twocolumn,letterpaper]{article}

\usepackage[pagenumbers]{cvpr} % To force page numbers, e.g. for an arXiv version

\usepackage[utf8]{inputenc} % allow utf-8 input
\usepackage{graphicx}
\usepackage{amsmath}
\usepackage{amssymb}
\usepackage{booktabs}
\usepackage{times}
\usepackage{epsfig} 
\usepackage{makecell}
\usepackage{times}
\usepackage{silence}
\usepackage{epstopdf}
\usepackage{color, colortbl}
\usepackage{float}  
\usepackage{caption}
\usepackage{multirow} 
\usepackage{xcolor}
\usepackage{url}
\usepackage{cite}
\usepackage{placeins}
\usepackage[group-separator={,}]{siunitx}
\usepackage[utf8]{inputenc}
\usepackage[pagebackref,breaklinks,colorlinks]{hyperref}
\usepackage{pifont}% http://ctan.org/pkg/pifont
\newcommand{\cmark}{\ding{51}}%
\newcommand{\xmark}{\ding{55}}%

\newcommand{\vspaceundertab}{\vspace{-.2cm}}
\newcommand{\vspaceunderfigext}{\vspace{-0.7cm}}
\newcommand{\vspaceundercaption}{\vspace{-0.3cm}}

\newcommand{\comment}[1]{}

\definecolor{LightCyan}{rgb}{0.88,1,1}
\definecolor{Gray}{gray}{0.9}
\definecolor{Pink}{rgb}{1, 0, 1}
\definecolor{azure}{rgb}{0.0, 0.44, 1.0}
\definecolor{bleudefrance}{rgb}{0.19, 0.55, 0.91}
\definecolor{cobalt}{rgb}{0.0, 0.28, 0.67}

\hypersetup{
  citecolor=cobalt,
}
\usepackage[pagebackref,breaklinks,colorlinks]{hyperref}

\newcommand{\re}[1]{\textcolor{red}{#1}}

\newcommand{\gr}[1]{\textcolor{gray}{#1}}
\newcommand{\bl}[1]{\textcolor{blue}{#1}}
\newcommand{\mg}[1]{\textcolor{magenta}{#1}}
\newcommand{\pp}[1]{\textcolor{purple}{#1}}

\newcommand{\az}[1]{\textcolor{azure}{#1}}
\newcommand{\dt}[1]{\fontsize{8pt}{0.1em}\selectfont (#1)}

\newlength\savewidth

\newcolumntype{x}[1]{>{\centering\arraybackslash}p{#1pt}}
\newcolumntype{y}[1]{>{\raggedright\arraybackslash}p{#1pt}}

\definecolor{LightCyan}{rgb}{0.88,1,1}
\definecolor{UIO}{RGB}{223,99,67}
\definecolor{UIB}{RGB}{24,41,72}
\definecolor{UOG}{RGB}{45,102,55}

\newcommand{\modelname}{OneFormer\xspace}

\usepackage[capitalize]{cleveref}
\crefname{section}{Sec.}{Secs.}
\Crefname{section}{Section}{Sections}
\Crefname{table}{Table}{Tables}
\crefname{table}{Tab.}{Tabs.}

\usepackage{txfonts}

\def\cvprPaperID{---} % *** Enter the CVPR Paper ID here
\def\confName{CVPR}
\def\confYear{---}

\begin{document}

\title{OneFormer: One Transformer to Rule Universal Image Segmentation}

\author{Jitesh Jain\textsuperscript{1,2}, Jiachen Li\textsuperscript{1$^*$}, MangTik Chiu\textsuperscript{1$^*$}, Ali Hassani\textsuperscript{1}, Nikita Orlov\textsuperscript{3}, Humphrey Shi\textsuperscript{1,3}\\
{\small \textsuperscript{1}SHI Labs @ U of Oregon \& UIUC, \textsuperscript{2}IIT Roorkee}, \small \textsuperscript{3}Picsart AI Research (PAIR)
\\
{\small \url{https://github.com/SHI-Labs/OneFormer}
}
}

\twocolumn[{%
\renewcommand\twocolumn[1][]{#1}%
\maketitle
\begin{center}
    \centering
    \captionsetup{type=figure}
    \vspace{-0.7cm}
    \includegraphics[width=\textwidth]{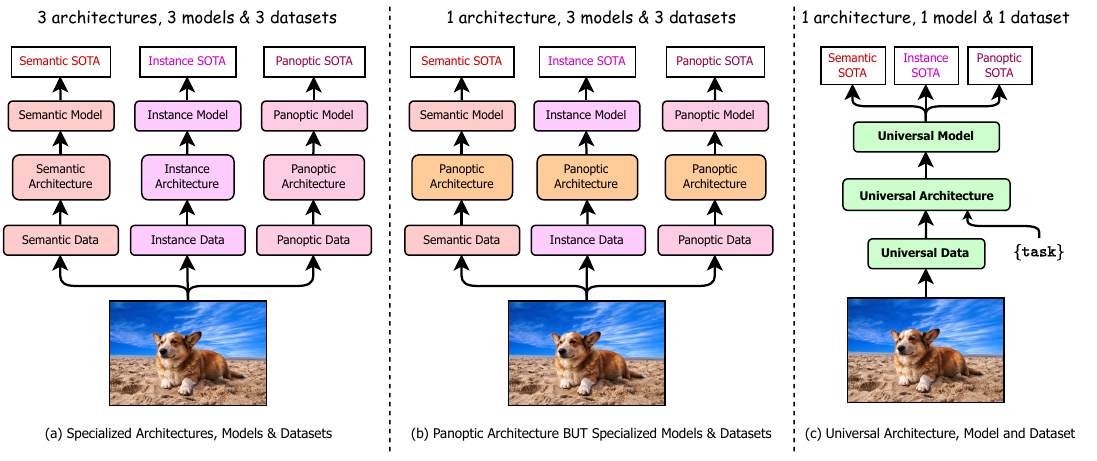}
    \vspace{-0.8cm}
    \captionof{figure}{\textbf{A Path to Universal Image Segmentation.} (a) Traditional segmentation methods developed specialized architectures and models for each task to achieve top performance. (b) Recently, new panoptic architectures~\cite{mask2former, knet} used the same architecture to achieve top performance across different tasks. However, they still need to train different models for different tasks, resulting in a semi-universal approach. (c) We propose a unique multi-task universal architecture with a task-conditioned joint training strategy that sets new state-of-the-arts across semantic, instance and panoptic segmentation tasks with a single model, unifying segmentation across architecture, model and dataset. Our work significantly reduces the underlying resource requirements and makes segmentation more universal and accessible.}
    \label{fig:comparison}
\end{center}%
}]

\begin{abstract}
\vspace{-0.45cm}
Universal Image Segmentation is not a new concept. Past attempts to unify image segmentation in the last decades include scene parsing, panoptic segmentation, and, more recently, new panoptic architectures. However, such panoptic architectures do not truly unify image segmentation because they need to be trained individually on the semantic, instance, or panoptic segmentation to achieve the best performance. Ideally, a truly universal framework should be trained only once and achieve SOTA performance across all three image segmentation tasks. To that end, we propose \modelname, a universal image segmentation framework that unifies segmentation with a multi-task train-once design. 
We first propose a task-conditioned joint training strategy that enables training on ground truths of each domain (semantic, instance, and panoptic segmentation) within a single multi-task training process. Secondly, we introduce a task token to condition our model on the task at hand, making our model task-dynamic to support multi-task training and inference.
Thirdly, we propose using a query-text contrastive loss during training to establish better inter-task and inter-class distinctions. Notably, our single \modelname model outperforms specialized Mask2Former models across all three segmentation tasks on ADE20k, Cityscapes, and COCO, despite the latter being trained on each of the three tasks individually with three times the resources. With new ConvNeXt and DiNAT backbones, we observe even more performance improvement. We believe \modelname is a significant step towards making image segmentation more universal and accessible. To support further research, we open-source our code and models at \href{https://github.com/SHI-Labs/OneFormer}{https://github.com/SHI-Labs/OneFormer}.

\end{abstract}
\vspace{-0.5cm}
\section{Introduction}

Image Segmentation is the task of grouping pixels into multiple segments. Such grouping can be semantic-based (\emph{e.g.}, road, sky, building), or instance-based (objects with well-defined boundaries). 
Earlier segmentation approaches~\cite{fcn, deeplabv2, mask-rcnn} tackled these two segmentation tasks individually, with specialized architectures and therefore separate research effort into each.
In a recent effort to unify semantic and instance segmentation, Kirillov \emph{et al.}~\cite{pq} proposed panoptic segmentation, with pixels grouped into an amorphous segment for amorphous background regions (labeled ``stuff'') and distinct segments for objects with well-defined shape (labeled ``thing''). This effort, however, led to new specialized panoptic architectures~\cite{panoptic-deeplab} instead of unifying the previous tasks (see ~\cref{fig:comparison}\re{a}).
More recently, the research trend shifted towards unifying image segmentation with new panoptic architectures, such as K-Net~\cite{knet}, MaskFormer~\cite{maskformer}, and Mask2Former~\cite{mask2former}. Such panoptic architectures can be trained on all three tasks and obtain high performance without changing architecture. They do need to, however, be trained individually on each task to achieve the best performance (see ~\cref{fig:comparison}\re{b}). The individual training policy requires extra training time and produces different sets of model weights for each task. In that regard, they can only be considered a semi-universal approach. For example, Mask2Former~\cite{mask2former} is trained for 160K iterations on ADE20K~\cite{ade20k} for each of the semantic, instance, and panoptic segmentation tasks to obtain the best performance for each task, yielding a total of 480k iterations in training, and three models to store and host for inference.

In an effort to truly unify image segmentation, we propose a multi-task universal image segmentation framework (\textbf{\modelname}), which outperforms existing state-of-the-arts on all three image segmentation tasks (see \cref{fig:comparison}\re{c}), by only training once on one panoptic dataset. Through this work, we aim to answer the following questions: 

\noindent (i) \textit{Why are existing panoptic architectures~\cite{mask2former, maskformer} not successful with a single training process or model to tackle all three tasks?} We hypothesize that existing methods need to train individually on each segmentation task due to the absence of task guidance in their architectures, making it challenging to learn the inter-task domain differences when trained jointly or with a single model. To tackle this challenge, we introduce a task input token in the form of text: ``the task is \{\texttt{task}\}", to condition the model on the task in focus, making our architecture task-guided for training, and task-dynamic for inference, all with a single model. We uniformly sample \{\texttt{task}\} from \texttt{\{panoptic, instance, semantic\}} and the corresponding ground truth during our joint training process to ensure our model is unbiased in terms of tasks. Motivated by the ability of panoptic~\cite{pq} data to capture both semantic and instance information, we derive the semantic and instance labels from the corresponding panoptic annotations during training. Consequently, we only need panoptic data during training. Moreover, our joint training time, model parameters, and FLOPs are comparable to the existing methods, decreasing training time and storage requirements up to 3\texttimes{}, making image segmentation less resource intensive and more accessible.

\smallskip
\noindent (ii) \textit{How can the multi-task model better learn inter-task and inter-class differences during the single joint training process?} Following the recent success of transformer frameworks~\cite{detr, semask, mask2former, kmax_deeplab, swin-T, nat, dinat} in computer vision, we formulate our framework as a transformer-based approach, which can be guided through the use of query tokens. To add task-specific context to our model, we initialize our queries as repetitions of the task token (obtained from the task input) and compute a query-text contrastive loss~\cite{groupvit, clip} with the text derived from the corresponding ground-truth label for the sampled task as shown in \cref{fig:cqformer}. We hypothesize that a contrastive loss on the queries helps guide the model to be more task-sensitive. Furthermore, it also helps reduce the category mispredictions to a certain extent. 
% We provide a more detailed overview of our framework in \cref{sec:method}.
% major image segmentation
% experiments demonstrate that 

We evaluate \modelname on three major segmentation datasets: ADE20K~\cite{ade20k}, Cityscapes~\cite{cityscapes}, and COCO~\cite{coco}, each with all three (semantic, instance, and panoptic) segmentation tasks. \modelname sets the new state of the arts for all three tasks with a single jointly trained model. To summarize, our main contributions are:
\vspace{-0.2cm}
\begin{itemize}
    % \item We propose a novel task-guided transformer-based framework \modelname, which outperforms existing frameworks by training only once for all three tasks, despite the latter being trained separately on each task. To the best of our knowledge, we are the first to propose a train-only-once universal image segmentation framework.
    \item We propose \modelname, the first multi-task universal image segmentation framework based on transformers that need to be trained only once with a single universal architecture, a single model, and on a single dataset, to outperform existing frameworks across semantic, instance, and panoptic segmentation tasks, despite the latter need to be trained separately on each task using multiple times of the resources.
    % a novel  framework, with  To the best of our knowledge, we are the first to propose a train-only-once universal image segmentation framework. 
    % that competes with the specialized state-of-the-arts on all three image segmentation tasks.
    \vspace{-0.15cm}
    \item OneFormer uses a task-conditioned joint training strategy, uniformly sampling different ground truth domains ( semantic, instance, or panoptic) by deriving all labels from panoptic annotations to train its multi-task model. Thus, OneFormer actually achieves the orignial unification goal of panoptic segmentation~\cite{pq}.
    
    \vspace{-0.15cm}
    \item We validate OneFormer through extensive experiments on three major benchmarks: ADE20K~\cite{ade20k}, Cityscapes~\cite{cityscapes}, and COCO~\cite{coco}. \modelname sets a new state-of-the-art performance on all three segmentation tasks compared with methods using the standard Swin-L~\cite{swin-T} backbone, and improves even more with new ConvNeXt~\cite{convnext} and DiNAT~\cite{dinat} backbones.
\end{itemize}

\section{Related Work}
\label{sec:rel_work}
\subsection{Image Segmentation}
Image segmentation is one of the most fundamental tasks in image processing and computer vision. Traditional works usually tackle one of the three image segmentation tasks with specialized network architectures (\cref{fig:comparison}\re{a}).
% : semantic, instance, and panoptic segmentation

\begin{figure*}[t!]
  \centering
\includegraphics[width=\linewidth]{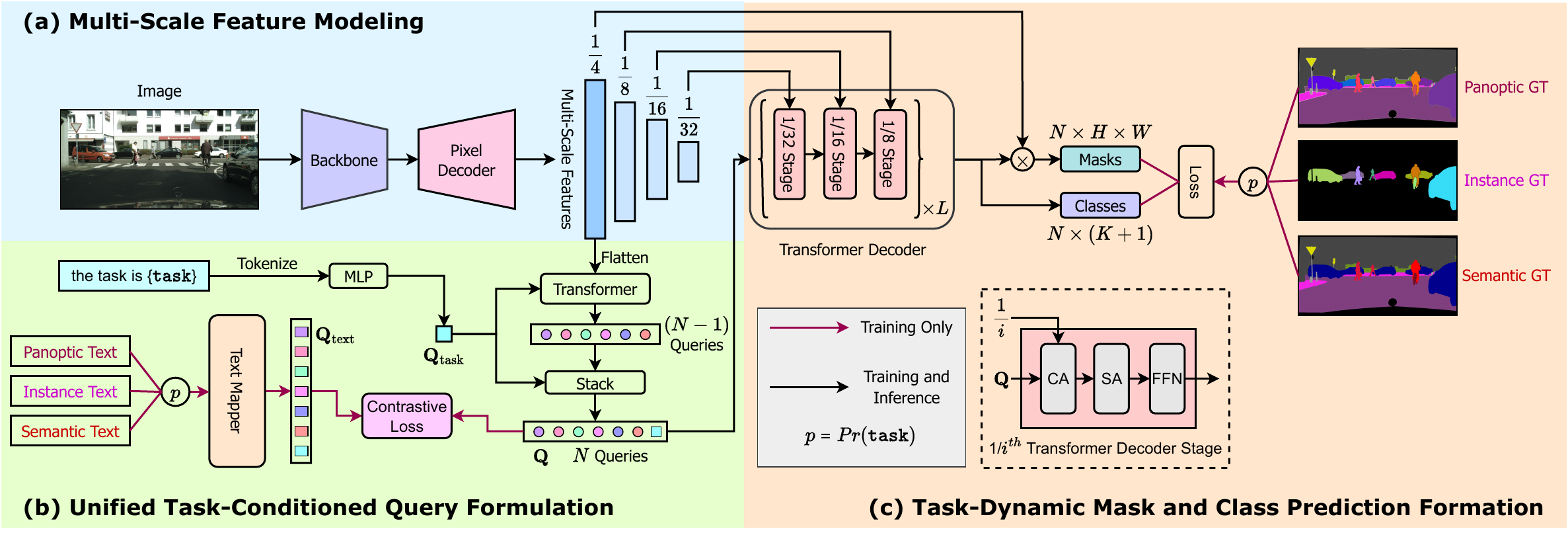}
  \vspaceunderfigext
  \caption{
      \textbf{\modelname Framework Architecture.} (a) We extract multi-scale features for an input image using a backbone, followed by a pixel decoder. (b) We formulate a unified set of $N-1$ task-conditioned object queries with guidance from the task token ($\mathbf{Q}_\text{task}$) and flattened $1/4$-scale features inside a transformer~\cite{vaswani2017attention}. Next, we concatenate $\mathbf{Q}_\text{task}$ with the $N-1$ queries from the transformer. We uniformly ($p=1/3$) sample the task during training and generate the corresponding text queries ($\mathbf{Q}_\text{text}$) using a text mapper (\cref{fig:text_mapper}). We calculate a query-text contrastive loss to learn the inter-task distinctions. We can drop the text mapper during inference, thus, making our model parameter efficient. (c) We use a multi-stage $L$-layer transformer decoder to obtain the task-dynamic class and mask predictions.
      }
  \label{fig:cqformer}
\vspaceundercaption
\end{figure*}
% an MSDeformAttn~\cite{deformable-detr} based 
% (panoptic, instance, or semantic)
% \textbf{Semantic Segmentation.} Semantic segmentation was originally formulated as a CNN-based per-pixel classification task in \cite{fcn}. Several following CNN-based works focus on capturing context by adding custom-designed convolutional modules~\cite{deeplabv1, deeplabv2, encnet, pspnet, segnet} or attention modules~\cite{ccnet, dualnet, attanet}. Many works~\cite{segmenter, segformer, semask, swin-T} have widely shown the success of vision transformer-based methods in semantic segmentation. 

\textbf{Semantic Segmentation.} Semantic segmentation was long tackled as a pixel classification problem with CNNs~\cite{fcn, deeplabv1, deeplabv2, cheng2019spgnet}. More recent works~\cite{ccnet, segmenter, segformer, semask} have shown the success of transformer-based methods in semantic segmentation following its success in language and vision~
\cite{vaswani2017attention,detr}. Among them, MaskFormer~\cite{maskformer} treated semantic segmentation as a mask classification problem following early works~\cite{hariharan2014simultaneous,carreira2012semantic,dai2015convolutional}, through using a transformer decoder with object queries~\cite{detr}. We also formulate semantic segmentation as a mask classification problem.
% 

% \textbf{\modelname Framework Architecture.} (a) We extract multi-scale features for an input image using a backbone, followed by an MSDeformAttn~\cite{deformable-detr} based pixel decoder. (b) We formulate a unified set of $N-1$ task-conditioned queries with guidance from the task token ($\mathbf{Q}_\text{task}$) and flattened $1/4$-scale features inside a transformer~\cite{vaswani2017attention}. Next, we concatenate $\mathbf{Q}_\text{task}$ with the $N-1$ queries from the transformer. We uniformly ($p=1/3$) sample the task (panoptic, instance, or semantic) during training and generate the corresponding text queries ($\mathbf{Q}_\text{text}$) using a text mapper (\cref{fig:text_mapper}). Then, we calculate a query-text contrastive loss to learn the task differences. We can drop the text mapper during inference, thus, making our model parameter efficient. (c) We use a multi-stage $L$-layer transformer decoder to obtain the task-dynamic class and mask predictions.
%       }
\textbf{Instance Segmentation.} Traditional instance segmentation methods~\cite{mask-rcnn, cai2018cascade, htc++} are also formulated as mask classifiers, which predict binary masks and a class label for each mask. We also formulate instance segmentation as a mask classification problem.

\textbf{Panoptic Segmentation.} Panoptic Segmentation~\cite{pq} was proposed to unify instance and semantic segmentation. One of the earliest architectures in this scope was Panoptic-FPN~\cite{sem-fpn}, which introduced separate instance and semantic task branches. Works that followed significantly improved performance with transformer-based architectures~\cite{max-deeplab, cmt_deeplab, kmax_deeplab, axial-deeplab, mask2former, maskformer}. Despite the progress made so far, panoptic segmentation models are still behind in performance compared to individual instance and semantic segmentation models, therefore not living up to their full unification potential. Motivated by this, we design our \modelname to be trained with panoptic annotations only.
% and outperforms or is on par with existing instance and semantic segmentation state-of-the-art methods with major improvements on panoptic segmentation.

\subsection{Universal Image Segmentation}

The concept of universal image segmentation has existed for some time, starting with image and scene parsing~\cite{describing_scene, image-parse, tighe2014scene}, followed by panoptic segmentation as an effort to unify semantic and instance segmentation~\cite{pq}. More recently, promising architectures~\cite{knet, mask2former, maskformer} designed specifically for panoptic segmentation have emerged which also perform well on semantic and instance segmentation tasks. K-Net~\cite{knet}, a CNN, uses dynamic learnable instance and semantic kernels with bipartite matching. MaskFormer~\cite{maskformer} is a transformer-based architecture, serving as a mask classifier. It was inspired by DETR's~\cite{detr} reformulation of object detection in the scope of transformers, where the image is fed to the encoder, and the decoder produces proposals based on queries. Mask2Former~\cite{mask2former} improved upon MaskFormer with learnable queries, deformable multi-scale attention~\cite{deformable-detr} in the decoder, a masked cross-attention and set the new state of the art on all three tasks. Unfortunately, it requires training the model individually on each task to achieve the best performance. Therefore, there remains a gap in truly unifying the three segmentation tasks. To the best of our knowledge, \modelname is the first framework to beat state of the art on all three image segmentation tasks with a single universal model.

\begin{figure*}[t!]
  \centering
\includegraphics[width=\linewidth]{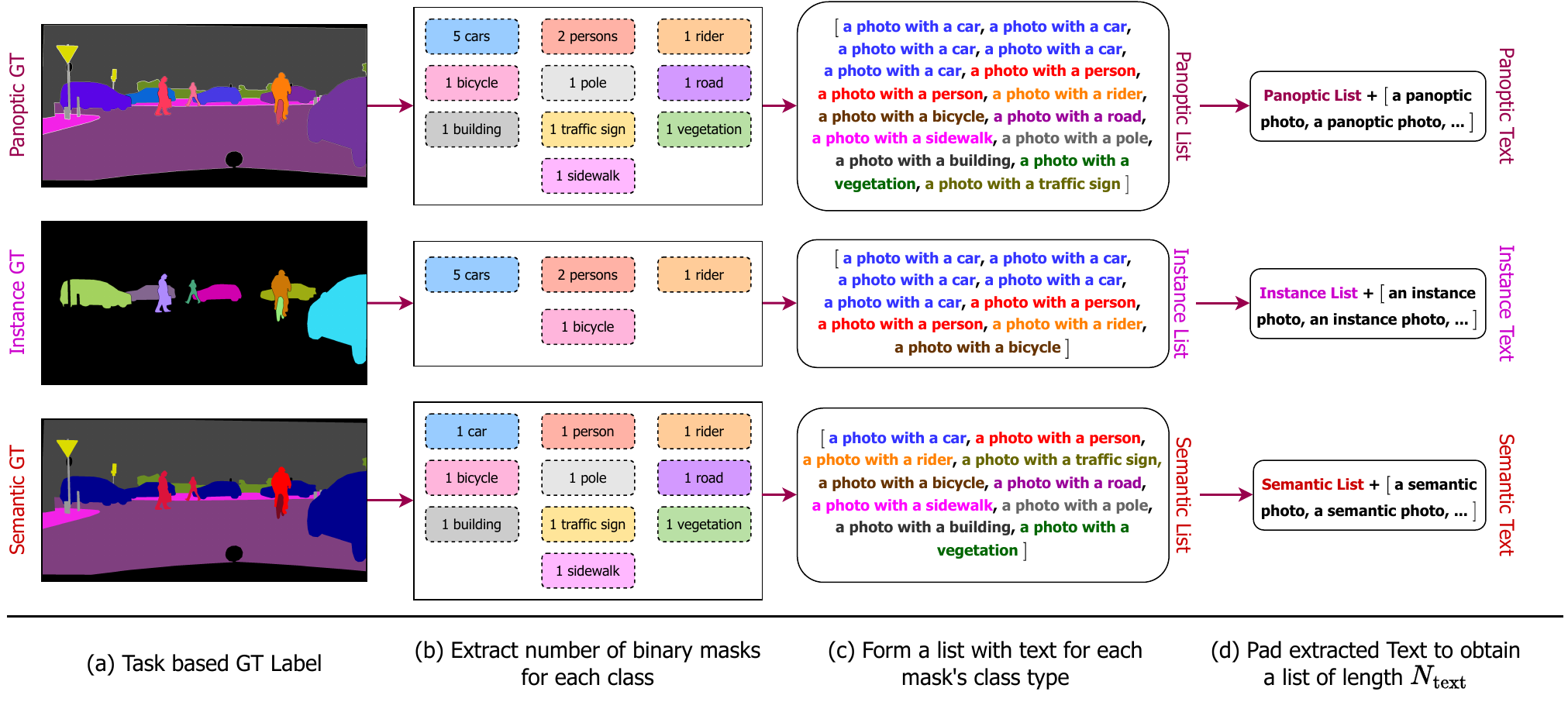}
  \vspaceunderfigext
  \caption{
      \textbf{Input Text Formation.} (a) We uniformly sample the \texttt{task} during training. (b) Following the task selection, we extract the number of distinct binary masks for each class to be detected from the corresponding GT label. (c) We form a list with text descriptions for each mask using the template ``a photo with a \{\texttt{CLS}\}", where \texttt{CLS} represents the corresponding class name for the object mask. (d) Finally, we pad the text list to a constant length of $N_\text{text}$ using ``a/an \{\texttt{task}\} photo" entries which represent the \texttt{no-object} detections; where \texttt{task} $\in$ \{panoptic, instance, semantic\}. 
      }
  \label{fig:text_gen}
\vspaceundercaption
\end{figure*}

\subsection{Transformer-based Architectures}
Architectures based on the transformer encoder-decoder structure~\cite{detr, deformable-detr, dn-detr, dabdetr} have proved effective in object detection since the introduction of DETR~\cite{detr}. Mask2Former~\cite{maskformer, mask2former} demonstrated the effectiveness of such architectures for image segmentation with a mask classification formulation. Inspired by this success, we also formulate our framework as a query-based mask classification task. Additionally, we claim that calculating a query-text contrastive loss~\cite{groupvit, clip} on the task-guided queries can help the model learn inter-task differences and reduce the category mispredictions in the model outputs. Concurrent to our work, LMSeg~\cite{lmseg} uses text derived from multiple datasets' taxonomy to calculate a query-text contrastive loss and tackle the multi-dataset segmentation training challenge. Unlike LMSeg~\cite{lmseg}, our work focuses on multiple tasks and uses the classes present in the training sample's ground-truth label to calculate the query-text contrastive loss.

% We provide a more detailed description of our method in the following section.

%  For smoother learning, we use a set of learnable context text queries~\cite{zhou2022coop, zhou2022cocoop} in addition to the mapped text input for matching the object queries to the text queries.
\section{Method}
\label{sec:method}

In this section, we introduce \modelname, a universal image segmentation framework jointly trained on the panoptic, semantic, and instance segmentation and outperforms individually trained models. We provide an overview of \modelname in \cref{fig:cqformer}. OneFormer uses two inputs: sample image and task input of the form ``the task is \{\texttt{task}\}". During our single joint training process, the \texttt{task} is uniformly sampled from \{panoptic, instance, semantic\} for each image. Firstly, we extract multi-scale features from the input image using a backbone and a pixel decoder. We tokenize the task input to obtain a 1-D task token used to condition the object queries and, consequently, our model on the task for each input. Additionally, we create a text list representing the number of binary masks for each class present in the GT label and map it to text query representations. Note that the text list depends on the input image and the \{\texttt{task}\}. For supervision of the model's task-dynamic predictions, we derive the corresponding ground-truths from panoptic annotations. As the ground truth is task-dependent, we calculate a query-text contrastive loss between the object and text queries to ensure there is task distinction in the object queries. The object queries and multi-scale features are fed into a transformer decoder to produce final predictions.  We provide more details in the following sections.

 % The text list is mapped to  that signify an image's ground truth segments.

\subsection{Task Conditioned Joint Training}
\label{subsec:joint}

Existing semi-universal architectures for image segmentation~\cite{knet, maskformer, mask2former} face a significant drop in performance when jointly trained on all three segmentation tasks (\cref{tab:ablat_joint}). We attribute their failure to tackle the multi-task challenge to the absence of task-conditioning in their architecture.

We tackle the multi-task train-once challenge for image segmentation using a task-conditioned joint training strategy. Particularly, we first uniformly sample the \texttt{task} from \{panoptic, semantic, instance\} for the GT label. We realize the unification potential of panoptic annotations~\cite{pq} by deriving the \texttt{task}-specific labels from the panoptic annotations, thus, using only one set of annotations. 
% \emph{i.e.}, when \texttt{task} is semantic/instance segmentation, we derive the corresponding label from the GT panoptic annotation for the sampled image.

Next, we extract a set of binary masks for each category present in the image from the task-specific GT label, \emph{i.e.}, semantic task guarantees only one amorphous binary mask for each class present in the image, whereas, instance task signifies non-overlapping binary masks for only thing classes, ignoring the stuff regions. Panoptic task denotes a single amorphous mask for stuff classes and non-overlapping masks for thing classes as shown in \cref{fig:text_gen}. Subsequently, we iterate over the set of masks to create a list of text ($\mathbf{T}_\text{list}$) with a template ``a photo with a \{\texttt{CLS}\}", where \texttt{CLS} is the class name for the corresponding binary mask. The number of binary masks per sample varies over the dataset. Therefore, we pad $\mathbf{T}_\text{list}$ with ``a/an \{\texttt{task}\} photo" entries to obtain a padded list ($\mathbf{T}_\text{pad}$) of constant length $N_\text{text}$, with padded entries representing \texttt{no-object} masks. We later use $\mathbf{T}_\text{pad}$ for computing a query-text contrastive loss (\cref{subsec:contra_q}).

We condition our architecture on the \texttt{task} using a task input ($\mathbf{I}_\text{task}$) with the template ``the task is \{\texttt{task}\}", which is tokenized and mapped to a task-token ($\mathbf{Q}_\text{task}$). We use $\mathbf{Q}_\text{task}$ to condition \modelname on the \texttt{task} (\cref{subsec:query}).

\subsection{Query Representations}
\label{subsec:query}

During training, we use two sets of queries in our architecture: text queries ($\mathbf{Q}_\text{text}$) and object queries ($\mathbf{Q}$). $\mathbf{Q}_\text{text}$ is the text-based representation for the segments in the image, while $\mathbf{Q}$ is the image-based representation.

\begin{figure}[t!]
  \centering
\includegraphics[width=\linewidth]{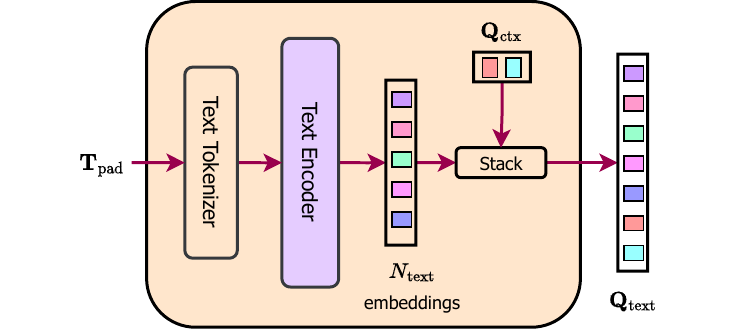}
  \caption{
      \textbf{Text Mapper.} We tokenize and then encode the input text list ($\mathbf{T}_\text{pad}$) using a 6-layer transformer text encoder~\cite{groupvit, vaswani2017attention} to obtain a set of $N_\text{text}$ embeddings. We concatenate a set of $N_\text{ctx}$ learnable embeddings to the encoded representations to obtain the final $N$ text queries ($\mathbf{Q}_\text{text}$). The $N$ text queries stand for a text-based representation of the objects present in an image.
      }
  \label{fig:text_mapper}
\vspaceundercaption
\end{figure}

To obtain $\mathbf{Q}_\text{text}$, we first tokenize the text entries $\mathbf{T}_\text{pad}$ and pass the tokenized representations through a text-encoder~\cite{groupvit}, which is a 6-layer transformer~\cite{vaswani2017attention}. The encoded $N_\text{text}$ text embeddings represent the number of binary masks and their corresponding classes in the input image. We further concatenate a set of $N_\text{ctx}$ learnable text context embeddings ($\mathbf{Q}_\text{ctx}$) to the encoded text embeddings to obtain the final $N$ text queries ($\mathbf{Q}_\text{text}$), as shown in \cref{fig:text_mapper}. Our motivation behind using $\mathbf{Q}_\text{ctx}$ is to learn a unified textual context~\cite{zhou2022coop, zhou2022cocoop} for a sample image. We only use the text queries during training; therefore, we can drop the text mapper module during inference to reduce the model size.

To obtain $\mathbf{Q}$, we first initialize the object queries ($\mathbf{Q'}$) as a $N-1$ times repetitions of the task-token ($\mathbf{Q}_\text{task}$). Then, we update $\mathbf{Q'}$ with guidance from the flattened $1/4$-scale features inside a 2-layer transformer~\cite{detr, vaswani2017attention}. The updated $\mathbf{Q'}$ from the transformer (rich with image-contextual information) is concatenated with $\mathbf{Q}_\text{task}$ to obtain a task-conditioned representation of $N$ queries, $\mathbf{Q}$. Unlike the vanilla all-zeros or random initialization~\cite{detr}, the task-guided initialization of the queries and the concatenation with $\mathbf{Q}_\text{task}$ is critical for the model to learn multiple segmentation tasks (\cref{subsec:ablations}).

% $^{\dag\dag}$: pretrained on COCO-Stuff-164k~\cite{coco-stuff}; $^*$: uses hidden dimension 1024 for the multi-scale pixel features.

\subsection{Task Guided Contrastive Queries}
\label{subsec:contra_q}

Developing a single model for all three segmentation tasks is challenging due to the inherent differences among the three tasks. The meaning of the object queries, $\mathbf{Q}$, is task-dependent. Should the queries focus only on the thing classes (instance segmentation), or should the queries predict only one amorphous object for each class present in the image (semantic segmentation) or a mix of both (panoptic segmentation)? Existing query-based architectures~\cite{maskformer, mask2former} do not take such differences into account and hence, fail at effectively training a single model on all three tasks.

To this end, we propose to calculate a query-text contrastive loss using  $\mathbf{Q}$ and $\mathbf{Q}_\text{text}$. We use $\mathbf{T}_\text{pad}$ to obtain the text queries representation, $\mathbf{Q}_{text}$, where $\mathbf{T}_{pad}$ is a list of textual representations for each mask-to-be-detected in a given image with ``a/an \{\texttt{task}\} photo" representing the \texttt{no-object} detections in  $\mathbf{Q}$~\cite{detr}. Thus, the text queries align with the purpose of object queries, representing the objects/segments present~\cite{detr} in an image. Therefore, we can successfully learn the inter-task distinctions in the query representations using a contrastive loss between the ground truth-derived text and object queries. Moreover, contrastive learning on the queries enables us to attend to inter-class differences and reduce category misclassifications. 
% for the objects' masks.

\vspace{-0.4cm}
\begin{equation}
    \begin{split}
        \mathcal{L}_{\mathbf{Q} \rightarrow {\mathbf{Q}_\text{text}}} & = -\frac{1}{B}\sum_{i=1}^B\log \frac{\exp(q_i^{obj} {\odot} q_i^{txt}/\tau)}{\sum_{j=1}^B \exp(q_i^{obj} {\odot} q_j^{txt}/\tau)}, \\
        \mathcal{L}_{{\mathbf{Q}_\text{text}} \rightarrow {\mathbf{Q}}} & = -\frac{1}{B}\sum_{i=1}^B\log \frac{\exp(q_i^{txt} {\odot} q_i^{obj}/\tau)}{\sum_{j=1}^B \exp(q_i^{txt} {\odot} q_j^{obj}/\tau)} \\
        \mathcal{L}_{\mathbf{Q} \leftrightarrow {\mathbf{Q}_\text{text}}} & = \mathcal{L}_{\mathbf{Q} \rightarrow {\mathbf{Q}_\text{text}}} + \mathcal{L}_{{\mathbf{Q}_\text{text}} \rightarrow {\mathbf{Q}}}
    \end{split}
    \label{eq:contra_loss}
\end{equation}

Considering that we have a batch of $B$ object-text query pairs $\{(q_i^{obj}, x_i^{txt})\}_{i=1}^B$, where $q_i^{obj}$ and $q_i^{txt}$ are the corresponding object and text queries, respectively, of the $i$-th pair, we measure the similarity between the queries by calculating a dot product. The total contrastive loss is composed of two losses~\cite{groupvit}: (i) an object-to-text contrastive loss ($\mathcal{L}_{\mathbf{Q} \rightarrow {\mathbf{Q}_\text{text}}}$) and; (ii) a text-to-object contrastive loss ($\mathcal{L}_{{\mathbf{Q}_\text{text}} \rightarrow {\mathbf{Q}}}$) as shown in \cref{eq:contra_loss}. $\tau$ is a learnable temperature parameter to scale the contrastive logits.

% \vspace{-0.4cm}
% \begin{equation}
%     \begin{split}
%         \mathcal{L}_{\mathbf{Q} \rightarrow {\mathbf{Q}_\text{text}}} & = -\frac{1}{B}\sum_{i=1}^B\log \frac{\exp(q_i^{obj} {\odot} q_i^{txt}/\tau)}{\sum_{j=1}^B \exp(q_i^{obj} {\odot} q_j^{txt}/\tau)}, \\
%         \mathcal{L}_{{\mathbf{Q}_\text{text}} \rightarrow {\mathbf{Q}}} & = -\frac{1}{B}\sum_{i=1}^B\log \frac{\exp(q_i^{txt} {\odot} q_i^{obj}/\tau)}{\sum_{j=1}^B \exp(q_i^{txt} {\odot} q_j^{obj}/\tau)} \\
%         \mathcal{L}_{\mathbf{Q} \leftrightarrow {\mathbf{Q}_\text{text}}} & = \mathcal{L}_{\mathbf{Q} \rightarrow {\mathbf{Q}_\text{text}}} + \mathcal{L}_{{\mathbf{Q}_\text{text}} \rightarrow {\mathbf{Q}}}
%     \end{split}
%     \label{eq:contra_loss}
% \end{equation}

% ; $^{\dag\dag}$: extra Mapillary Vistas~\cite{mapillary} dataset training; $^\ddag$: uses hidden dimension 1024 for the multi-scale pixel features. The rows with \gr{gray} text denote evaluation on multi-scale inputs

% , $^*$ hidden dimension 1024, $^{\dag\dag}$: uses extra training data. Text in \gr{gray} denotes sliding fixed size inference.

\begin{table*}[t!]\setlength{\tabcolsep}{10pt}
  \centering
  \resizebox{1.\linewidth}{!}{
  \begin{tabular}{ll|ccc|cc|cccc}
   \textbf{Method} & \textbf{Backbone} & \#\textbf{Params} & \#\textbf{FLOPs} & \#\textbf{Queries} & \textbf{Crop Size} & \begin{tabular}{@{}c@{}} \textbf{Iters} \end{tabular} & \textbf{PQ} & \textbf{AP} & \begin{tabular}{@{}c@{}} \textbf{mIoU} \\ (s.s.) \end{tabular} & \begin{tabular}{@{}c@{}} \textbf{mIoU} \\ (m.s.) \end{tabular} \\

    \midrule
    \midrule
    \multicolumn{9}{l}{\textit{\textbf{Individual Training}}}\\
    \midrule
    \midrule
    
    \begin{tabular}{@{}l@{}}UPerNet$^\ddag$~\cite{upernet} \end{tabular} & SwinV2-L$^{\dag}$~\cite{swinv2} & --- & --- & --- & $640\!\times\!640$ & 40k & \begin{tabular}{@{}c@{}} --- \end{tabular} & \begin{tabular}{@{}c@{}} --- \end{tabular} & \begin{tabular}{@{}c@{}} ---- \end{tabular} & 55.9 \\
    
    \begin{tabular}{@{}l@{}}SeMask Mask2Former~\cite{semask} \end{tabular} & SeMask Swin-L$^{\dag}$~\cite{semask} & 223M & 426G & 200 & $640\!\times\!640$ & 160k & \begin{tabular}{@{}c@{}} --- \end{tabular} & \begin{tabular}{@{}c@{}} --- \end{tabular} & \begin{tabular}{@{}c@{}} 56.4 \end{tabular} & 57.5 \\
    \midrule
    
     \begin{tabular}{@{}l@{}}UPerNet + K-Net~\cite{knet} \end{tabular} & Swin-L$^\dag$~\cite{swin-T} & --- & --- & --- & $640\!\times\!640$ & 160k & \begin{tabular}{@{}c@{}} --- \end{tabular} & \begin{tabular}{@{}c@{}} --- \end{tabular} & \begin{tabular}{@{}c@{}} --- \end{tabular} & 54.3 \\
    
    \begin{tabular}{@{}l@{}}MaskFormer~\cite{maskformer} \end{tabular} & Swin-L$^\dag$~\cite{swin-T} & 212M & 375G & 100 & $640\!\times\!640$ & 160k & \begin{tabular}{@{}c@{}} --- \end{tabular} & \begin{tabular}{@{}c@{}} --- \end{tabular} & \begin{tabular}{@{}c@{}} 54.1 \end{tabular} & 55.6 \\
    
    \begin{tabular}{@{}l@{}}Mask2Former-Panoptic$^*$~\cite{mask2former} \end{tabular} & Swin-L$^\dag$~\cite{swin-T} & 216M & 413G & 200 & $640\!\times\!640$ & 160k  & \begin{tabular}{@{}c@{}} 48.7 \end{tabular} & \begin{tabular}{@{}c@{}} 34.2 \end{tabular} & \begin{tabular}{@{}c@{}} 54.5 \end{tabular} & \begin{tabular}{@{}c@{}} --- \end{tabular} \\
    
    \begin{tabular}{@{}l@{}}Mask2Former-Instance~\cite{mask2former} \end{tabular} & Swin-L$^\dag$~\cite{swin-T} & 216M & 411G & 200 & $640\!\times\!640$ & 160k  & \begin{tabular}{@{}c@{}} --- \end{tabular} & \begin{tabular}{@{}c@{}} 34.9 \end{tabular} & \begin{tabular}{@{}c@{}} --- \end{tabular} & \begin{tabular}{@{}c@{}} --- \end{tabular} \\
    
    \begin{tabular}{@{}l@{}}Mask2Former-Semantic~\cite{mask2former} \end{tabular} & Swin-L$^\dag$~\cite{swin-T} & 215M & 403G & 100 & $640\!\times\!640$ & 160k  & \begin{tabular}{@{}c@{}} --- \end{tabular} & \begin{tabular}{@{}c@{}} --- \end{tabular} & \begin{tabular}{@{}c@{}} 56.1 \end{tabular} & \begin{tabular}{@{}c@{}} 57.3 \end{tabular} \\

    \midrule
    \begin{tabular}{@{}l@{}}kMaX-DeepLab$^{\ddag\ddag}$~\cite{kmax_deeplab} \end{tabular} & ConvNeXt-L$^\dag$~\cite{convnext} & 232M & 333G & 256 & $641\!\times\!641$ & 100k & \begin{tabular}{@{}c@{}} 48.7 \end{tabular} & \begin{tabular}{@{}c@{}} --- \end{tabular} & \begin{tabular}{@{}c@{}} 54.8 \end{tabular} & --- \\

    \begin{tabular}{@{}l@{}}kMaX-DeepLab$^{\ddag\ddag}$~\cite{kmax_deeplab} \end{tabular} & ConvNeXt-L$^\dag$~\cite{convnext} & 232M & 1302G & 256 & $1281\!\times\!1281$ & 100k & \begin{tabular}{@{}c@{}} 50.9 \end{tabular} & \begin{tabular}{@{}c@{}} --- \end{tabular} & \begin{tabular}{@{}c@{}} 55.2 \end{tabular} & --- \\
    
    \midrule
    \begin{tabular}{@{}l@{}}UPerNet$^{\ddag\ddag}$~\cite{upernet} \end{tabular} & SwinV2-G$^{{\dag\dag}}$~\cite{swinv2} & $>$3B & --- &  --- & $640\!\times\!640$ & 80k & \begin{tabular}{@{}c@{}} --- \end{tabular} & \begin{tabular}{@{}c@{}} --- \end{tabular} & \begin{tabular}{@{}c@{}} 59.1 \end{tabular} & --- \\
    
    % \midrule
    \begin{tabular}{@{}l@{}}Mask2Former$^{\ddag\ddag}$~\cite{mask2former} \end{tabular} & BEiT-3$^{{\dag\dag}}$~\cite{beit-3} & 1.9B & --- & --- & $896\!\times\!896$ & --- & \begin{tabular}{@{}c@{}} --- \end{tabular} & \begin{tabular}{@{}c@{}} --- \end{tabular} & \begin{tabular}{@{}c@{}} 62.0 \end{tabular} & 62.8 \\

    \midrule
    \midrule
    \multicolumn{9}{l}{\textit{\textbf{Joint Training}}}\\
    \midrule
    \midrule
    
    \begin{tabular}{@{}l@{}}\textbf{\modelname} \end{tabular} & Swin-L$^\dag$~\cite{swin-T} & 219M & 436G & 250 & $640\!\times\!640$ & 160k  & \begin{tabular}{@{}c@{}} \textbf{49.8} \end{tabular} & \begin{tabular}{@{}c@{}} \textbf{35.9} \end{tabular} & \begin{tabular}{@{}c@{}} \textbf{57.0} \end{tabular} & \begin{tabular}{@{}c@{}} \textbf{57.7} \end{tabular} \\
    
    \begin{tabular}{@{}l@{}}\textbf{\modelname} \end{tabular} & Swin-L$^\dag$~\cite{swin-T} & 219M & 801G & 250 & $896\!\times\!896$ & 160k  & \begin{tabular}{@{}c@{}} \textbf{51.1} \end{tabular} & \begin{tabular}{@{}c@{}} \textbf{37.6} \end{tabular} & \begin{tabular}{@{}c@{}} \textbf{57.4} \end{tabular} & \begin{tabular}{@{}c@{}} \textbf{58.3} \end{tabular} \\

    \begin{tabular}{@{}l@{}}\textbf{\modelname} \end{tabular} & Swin-L$^\dag$~\cite{swin-T} & 219M & 1597G & 250 & $1280\!\times\!1280$ & 160k  & \begin{tabular}{@{}c@{}} \textbf{51.4} \end{tabular} & \begin{tabular}{@{}c@{}} \textbf{37.8} \end{tabular} & \begin{tabular}{@{}c@{}} \textbf{57.0} \end{tabular} & \begin{tabular}{@{}c@{}} \textbf{57.7} \end{tabular} \\
    
    \midrule
    \begin{tabular}{@{}l@{}}\textbf{\modelname} \end{tabular} & ConvNeXt-L$^\dag$~\cite{convnext} & 220M & 389G & 250 & $640\!\times\!640$ & 160k  & \begin{tabular}{@{}c@{}} \textbf{50.0} \end{tabular} & \begin{tabular}{@{}c@{}} \textbf{36.2} \end{tabular} & \begin{tabular}{@{}c@{}} \textbf{56.6} \end{tabular} & \begin{tabular}{@{}c@{}} \textbf{57.4} \end{tabular} \\
    
    \begin{tabular}{@{}l@{}}\textbf{\modelname} \end{tabular} & ConvNeXt-XL$^\dag$~\cite{convnext} & 372M & 607G & 250 & $640\!\times\!640$ & 160k  & \begin{tabular}{@{}c@{}} \textbf{50.1} \end{tabular} & \begin{tabular}{@{}c@{}} \textbf{36.3} \end{tabular} & \begin{tabular}{@{}c@{}} \textbf{57.4} \end{tabular} & \begin{tabular}{@{}c@{}} \textbf{58.8} \end{tabular} \\
    
    \midrule
    \begin{tabular}{@{}l@{}}\textbf{\modelname} \end{tabular} & DiNAT-L$^\dag$~\cite{dinat} & 223M & 359G & 250 & $640\!\times\!640$ & 160k  & \begin{tabular}{@{}c@{}} \textbf{50.5} \end{tabular} & \begin{tabular}{@{}c@{}} \textbf{36.0} \end{tabular} & \begin{tabular}{@{}c@{}} \textbf{58.3} \end{tabular} & \begin{tabular}{@{}c@{}} \textbf{58.4} \end{tabular} \\
    
    \begin{tabular}{@{}l@{}}\textbf{\modelname} \end{tabular} & DiNAT-L$^\dag$~\cite{dinat} & 223M & 678G & 250 & $896\!\times\!896$ & 160k  & \begin{tabular}{@{}c@{}} \textbf{51.2} \end{tabular} & \begin{tabular}{@{}c@{}} \textbf{36.8} \end{tabular} & \begin{tabular}{@{}c@{}} \textbf{58.1} \end{tabular} & \begin{tabular}{@{}c@{}} \textbf{58.6} \end{tabular} \\

    \begin{tabular}{@{}l@{}}\textbf{\modelname} \end{tabular} & DiNAT-L$^\dag$~\cite{dinat} & 223M & 1369G & 250 & $1280\!\times\!1280$ & 160k  & \begin{tabular}{@{}c@{}} \textbf{51.5} \end{tabular} & \begin{tabular}{@{}c@{}} \textbf{37.1} \end{tabular} & \begin{tabular}{@{}c@{}} \textbf{58.2} \end{tabular} & \begin{tabular}{@{}c@{}} \textbf{58.7} \end{tabular} \\
    
    \bottomrule
    
\end{tabular}}
  \vspaceundertab
  \caption{\textbf{SOTA Comparison on the ADE20K val set.} $^\dag$: backbones pretrained on ImageNet-22K, $^*$: 0.5 confidence threshold; $^\ddag$: trained with batch size 32, $^{\ddag\ddag}$: trained with batch size 64. \modelname outperforms the individually trained Mask2Former~\cite{mask2former} on all metrics. Mask2Former's performance with 250 queries is not listed, as its performance degrades with 250 queries. We compute FLOPs using the corresponding crop size.
      }
    \label{tab:sota_ade20k}
\vspaceundercaption
\end{table*}

\subsection{Other Architecture Components}

\noindent
\textbf{Backbone and Pixel Decoder}: We use the widely used ImageNet~\cite{imagenet} pre-trained backbones~\cite{swin-T, convnext, dinat} to extract multi-scale feature representations from the input image. Our pixel decoder aids the feature modeling by gradually upsampling the backbone features. Motivated by the recent success of multi-scale deformable attention~\cite{deformable-detr, mask2former}, we use the same Multi-Scale Deformable Transformer (MSDeformAttn) based architecture for our pixel decoder.

\noindent
\textbf{Transformer Decoder}:  We use a multi-scale strategy~\cite{mask2former} to utilize the higher resolution maps inside our transformer decoder. Specifically, we feed the object queries ($\mathbf{Q}$) and the multi-scale outputs from the pixel decoder ($F_i$), $i \in \{1/4, 1/8, 1/16, 1/32\}$ as inputs. We use the features with resolution $1/8$, $1/16$ and $1/32$ of the original image alternatively to update $\mathbf{Q}$ using a masked cross-attention (CA) operation~\cite{mask2former}, followed by a self-attention (SA) and finally a feed-forward network (FFN). We perform these sets of alternate operations $L$ times inside the transformer decoder.

% on $\mathbf{Q}$

The final query outputs from the transformer decoder are mapped to a $K+1$ dimensional space for class predictions, where $K$ denotes the number of classes and an extra $+1$ for the \texttt{no-object} predictions. To obtain the final masks, we decode the pixel features ($F_{1/4}$) at $1/4$ resolution of the original image with the help of an \texttt{einsum} operation between $Q$ and $F_{1/4}$. During inference, we follow the same post-processing technique as \cite{mask2former} to obtain the final panoptic, semantic, and instance segmentation predictions. We only keep predictions with scores above a threshold of 0.5, 0.8, and 0.8 during post-processing for panoptic segmentation on the ADE20K~\cite{ade20k}, Cityscapes~\cite{cityscapes} and COCO~\cite{coco} datasets, respectively.

\subsection{Losses}

In addition to the contrastive loss on the queries, we calculate the standard classification CE-loss ($\mathcal{L}_\text{cls}$) over the class predictions. Following ~\cite{mask2former}, we use a combination of binary cross-entropy ($\mathcal{L}_\text{bce}$) and dice loss ($\mathcal{L}_\text{dice}$) over the mask predictions. Therefore, our final loss function is a weighted sum of the four losses (\cref{eq:loss}). We empirically set $\lambda_{\mathbf{Q} \leftrightarrow {\mathbf{Q}_\text{text}}}=0.5$, $\lambda_\text{cls}=2$, $\lambda_\text{bce}=5$ and $\lambda_\text{dice}=5$. To find the least cost assignment, we use bipartite matching~\cite{detr, maskformer} between the set predictions and the ground truths. We set $\lambda_\text{cls}$ as $0.1$ for the \texttt{no-object} predictions~\cite{mask2former}.

\vspace{-0.5cm}
\begin{equation}
\begin{split}
    \mathcal{L}_\text{final} & = \lambda_{\mathbf{Q} \leftrightarrow {\mathbf{Q}_\text{text}}}\mathcal{L}_{\mathbf{Q} \leftrightarrow {\mathbf{Q}_\text{text}}} + \lambda_\text{cls}\mathcal{L}_{cls} \\ & + \lambda_\text{bce}\mathcal{L}_{bce} + \lambda_\text{dice}\mathcal{L}_{dice}
\end{split}
\label{eq:loss}
\end{equation}
\section{Experiments}
\label{sec:experiments}

We illustrate that \modelname, when trained only once with our task-conditioned joint-training strategy, generalizes well to all three image segmentation tasks on three widely used datasets. Furthermore, we provide extensive ablations to demonstrate the significance of \modelname's components. Due to space constraints, we provide implementation details in the appendix.

\begin{table*}[t!]\setlength{\tabcolsep}{10pt}
  \centering
  \resizebox{1.\linewidth}{!}{
  \begin{tabular}{ll|ccc|cc|cccc}
   \textbf{Method} & \textbf{Backbone} & \#\textbf{Params} & \#\textbf{FLOPs} & \#\textbf{Queries} & \textbf{Crop Size} & \begin{tabular}{@{}c@{}} \textbf{Iters} \end{tabular} & \textbf{PQ} & \textbf{AP} & \begin{tabular}{@{}c@{}} \textbf{mIoU} \\ (s.s.) \end{tabular} & \begin{tabular}{@{}c@{}} \textbf{mIoU} \\ (m.s.) \end{tabular} \\

    \midrule
    \midrule
    \multicolumn{9}{l}{\textit{\textbf{Individual Training}}}\\
    \midrule
    \midrule
    
    \begin{tabular}{@{}l@{}}CMT-DeepLab$^\ddag$~\cite{cmt_deeplab} \end{tabular} & MaX-S$^\dag$~\cite{max-deeplab} & --- & --- & --- & $1025\!\times\!2049$ & 60k & \begin{tabular}{@{}c@{}} 64.6 \end{tabular} & \begin{tabular}{@{}c@{}} --- \end{tabular} & \begin{tabular}{@{}c@{}} 81.4 \end{tabular} & --- \\
    
    \begin{tabular}{@{}l@{}}Axial-DeepLab-L$^\ddag$~\cite{axial-deeplab} \end{tabular} & Axial ResNet-L$^\dag$~\cite{axial-deeplab} & 45M & 687G & --- & $1025\!\times\!2049$ & 60k & \begin{tabular}{@{}c@{}} 63.9 \end{tabular} & \begin{tabular}{@{}c@{}} 35.8 \end{tabular} & \begin{tabular}{@{}c@{}} 81.0 \end{tabular} & 81.5 \\
    
    \begin{tabular}{@{}l@{}}Axial-DeepLab-XL$^\ddag$~\cite{axial-deeplab} \end{tabular} & Axial ResNet-XL$^\dag$~\cite{axial-deeplab} & 173M & 2447G & --- & $1025\!\times\!2049$ & 60k & \begin{tabular}{@{}c@{}} 64.4 \end{tabular} & \begin{tabular}{@{}c@{}} 36.7 \end{tabular} & \begin{tabular}{@{}c@{}} 80.6 \end{tabular} & 81.1 \\

    \begin{tabular}{@{}l@{}}Panoptic-DeepLab$^\ddag$~\cite{panoptic-deeplab} \end{tabular} & SWideRNet$^{\dag}$~\cite{swidernet_2020} & 536M & 10365G & --- & $1025\!\times\!2049$ & 60k & \begin{tabular}{@{}c@{}} 66.4 \end{tabular} & \begin{tabular}{@{}c@{}} 40.1 \end{tabular} & \begin{tabular}{@{}c@{}} 82.2 \end{tabular} & 82.9 \\

    \midrule
    \begin{tabular}{@{}l@{}}Mask2Former-Panoptic~\cite{mask2former} \end{tabular} & Swin-L$^\dag$~\cite{swin-T} & 216M & 514G & 200 & $512\!\times\!1024$ & 90k  & \begin{tabular}{@{}c@{}} 66.6  \end{tabular} & \begin{tabular}{@{}c@{}} 43.6 \end{tabular} & \begin{tabular}{@{}c@{}} 82.9 \end{tabular} & \begin{tabular}{@{}c@{}} --- \end{tabular} \\
    
    \begin{tabular}{@{}l@{}}Mask2Former-Instance~\cite{mask2former} \end{tabular} & Swin-L$^\dag$~\cite{swin-T} & 216M & 507G & 200 & $512\!\times\!1024$ & 90k  & \begin{tabular}{@{}c@{}} ---  \end{tabular} & \begin{tabular}{@{}c@{}} 43.7 \end{tabular} & \begin{tabular}{@{}c@{}} --- \end{tabular} & \begin{tabular}{@{}c@{}} --- \end{tabular} \\
    
    \begin{tabular}{@{}l@{}}Mask2Former-Semantic~\cite{mask2former} \end{tabular} & Swin-L$^\dag$~\cite{swin-T} & 215M & 494G & 100 & $512\!\times\!1024$ & 90k  & \begin{tabular}{@{}c@{}} ---  \end{tabular} & \begin{tabular}{@{}c@{}} --- \end{tabular} & \begin{tabular}{@{}c@{}} 83.3 \end{tabular} & \begin{tabular}{@{}c@{}} 84.3 \end{tabular} \\
    
    \midrule
    \begin{tabular}{@{}l@{}}kMaX-DeepLab$^\ddag$~\cite{kmax_deeplab} \end{tabular} & ConvNeXt-L$^\dag$~\cite{convnext} & 232M & 1673G & 256 & $1025\!\times\!2049$ & 60k & \begin{tabular}{@{}c@{}} 68.4 \end{tabular} & \begin{tabular}{@{}c@{}} 44.0 \end{tabular} & \begin{tabular}{@{}c@{}} 83.5 \end{tabular} & --- \\

    \midrule
    \midrule
    \multicolumn{9}{l}{\textit{\textbf{Joint Training}}}\\
    \midrule
    \midrule
    
    \begin{tabular}{@{}l@{}}\textbf{\modelname} \end{tabular} & Swin-L$^\dag$~\cite{swin-T} & 219M & 543G & 250 & $512\!\times\!1024$ & 90k  & \begin{tabular}{@{}c@{}} \textbf{67.2} \end{tabular} & \begin{tabular}{@{}c@{}} \textbf{45.6} \end{tabular} & \begin{tabular}{@{}c@{}} 83.0 \end{tabular} & \begin{tabular}{@{}c@{}} \textbf{84.4} \end{tabular} \\
    
    \midrule
    \begin{tabular}{@{}l@{}}\textbf{\modelname} \end{tabular} & ConvNeXt-L$^\dag$~\cite{convnext} & 220M & 497G & 250 & $512\!\times\!1024$ & 90k  & \begin{tabular}{@{}c@{}} \textbf{68.5} \end{tabular} & \begin{tabular}{@{}c@{}} \textbf{46.5} \end{tabular} & \begin{tabular}{@{}c@{}} 83.0 \end{tabular} & \begin{tabular}{@{}c@{}} 84.0 \end{tabular} \\
    
    \begin{tabular}{@{}l@{}}\textbf{\modelname} \end{tabular} & ConvNeXt-XL$^\dag$~\cite{convnext} & 372M & 775G & 250 & $512\!\times\!1024$ & 90k  & \begin{tabular}{@{}c@{}} \textbf{68.4} \end{tabular} & \begin{tabular}{@{}c@{}} \textbf{46.7} \end{tabular} & \begin{tabular}{@{}c@{}} \textbf{83.6} \end{tabular} & \begin{tabular}{@{}c@{}} \textbf{84.6} \end{tabular} \\
    
    \midrule
    \begin{tabular}{@{}l@{}}\textbf{\modelname} \end{tabular} & DiNAT-L$^\dag$~\cite{dinat} & 223M & 450G & 250 & $512\!\times\!1024$ & 90k  & \begin{tabular}{@{}c@{}} \textbf{67.6} \end{tabular} & \begin{tabular}{@{}c@{}} \textbf{45.6} \end{tabular} & \begin{tabular}{@{}c@{}} 83.1 \end{tabular} & \begin{tabular}{@{}c@{}} 84.0 \end{tabular} \\
    
    \bottomrule
    
\end{tabular}}
  \vspaceundertab
  \caption{\textbf{SOTA Comparison on Cityscapes val set.} $^\dag$: backbones pretrained on ImageNet-22K; $^\ddag$: trained with batch size 32, $^*$: hidden dimension 1024. \modelname outperforms the individually trained Mask2Former~\cite{mask2former} models. Mask2Former's performance with 250 queries is not listed, as its performance degrades with 250 queries.  We compute FLOPs using the corresponding crop size.
      }

    \label{tab:sota_city}
\vspaceundercaption
\end{table*}

\begin{table*}[t!]\setlength{\tabcolsep}{10pt}
  \centering
  \resizebox{1.\linewidth}{!}{
  \begin{tabular}{ll|cccc|ccc|cc|c}
  \textbf{Method} & \textbf{Backbone} & \#\textbf{Params} & \#\textbf{FLOPs} & \#\textbf{Queries} & \begin{tabular}{@{}c@{}} \textbf{Epochs} \end{tabular} & \textbf{PQ} & \textbf{PQ}$^\text{Th}$ & \textbf{PQ}$^\text{St}$ & \textbf{AP} & \gr{\textbf{AP}$^\text{instance}$} & \begin{tabular}{@{}c@{}} \textbf{mIoU} \end{tabular} \\

   \midrule
    \midrule
    \multicolumn{9}{l}{\textit{\textbf{Individual Training}}}\\
    \midrule
    \midrule
    
    \begin{tabular}{@{}l@{}} MaskFormer~\cite{maskformer} \end{tabular} & Swin-L$^\dag$~\cite{swin-T} & 212M & 792G & 100 & 300  & 52.7 & \begin{tabular}{@{}c@{}} 58.5 \end{tabular} & \begin{tabular}{@{}c@{}} 44.0 \end{tabular} & --- & \gr{---} & 64.8 \\
    
    \begin{tabular}{@{}l@{}} K-Net~\cite{knet} \end{tabular} & Swin-L$^\dag$~\cite{swin-T} & --- & --- & 100 & 36 & 54.6 & \begin{tabular}{@{}c@{}} 60.2 \end{tabular} & \begin{tabular}{@{}c@{}} 46.0 \end{tabular} & --- & \gr{---} & --- \\
    
    \begin{tabular}{@{}l@{}} Panoptic SegFormer~\cite{panoptic-segformer} \end{tabular} & Swin-L$^\dag$~\cite{swin-T} & 221M & 816G & 353 & 24  & 55.8 & \begin{tabular}{@{}c@{}} 61.7 \end{tabular} & \begin{tabular}{@{}c@{}} 46.9 \end{tabular} & --- & \gr{---} & --- \\

    \begin{tabular}{@{}l@{}}Mask2Former-Panoptic~\cite{mask2former} \end{tabular} & Swin-L$^\dag$~\cite{swin-T} & 216M & 875G & 200 & 100  & \begin{tabular}{@{}c@{}} 57.8 \end{tabular} & \begin{tabular}{@{}c@{}} 64.2  \end{tabular} & \begin{tabular}{@{}c@{}} \textbf{48.1}
    \end{tabular} & \begin{tabular}{@{}c@{}} 48.7 \end{tabular} & \gr{48.6} & \begin{tabular}{@{}c@{}} \textbf{67.4} \end{tabular} \\
    
    \begin{tabular}{@{}l@{}}Mask2Former-Instance~\cite{mask2former} \end{tabular} & Swin-L$^\dag$~\cite{swin-T} & 216M & 868G & 200 & 100  & \begin{tabular}{@{}c@{}} --- \end{tabular} & \begin{tabular}{@{}c@{}} ---  \end{tabular} & \begin{tabular}{@{}c@{}} ---
    \end{tabular} & \begin{tabular}{@{}c@{}} \textbf{49.1} \end{tabular} & \gr{50.1} & \begin{tabular}{@{}c@{}} --- \end{tabular} \\
    
    \begin{tabular}{@{}l@{}}Mask2Former-Semantic$^\ddag$~\cite{mask2former} \end{tabular} & Swin-L$^\dag$~\cite{swin-T} & 216M & 891G & 200 & 100  & \begin{tabular}{@{}c@{}} --- \end{tabular} & \begin{tabular}{@{}c@{}} ---  \end{tabular} & \begin{tabular}{@{}c@{}} ---
    \end{tabular} & \begin{tabular}{@{}c@{}} --- \end{tabular} & \gr{---} & \begin{tabular}{@{}c@{}} 67.2 \end{tabular} \\
    
    \midrule
    \begin{tabular}{@{}l@{}}kMaX-DeepLab$^*$~\cite{kmax_deeplab} \end{tabular} & ConvNeXt-L$^\dag$~\cite{convnext} & 232M & 749G & 128 & 81  & \textbf{57.9} & \begin{tabular}{@{}c@{}} 64.0 \end{tabular} & \begin{tabular}{@{}c@{}} \textbf{48.6} \end{tabular} & --- & \gr{---} & --- \\

    \begin{tabular}{@{}l@{}}kMaX-DeepLab$^*$~\cite{kmax_deeplab} \end{tabular} & ConvNeXt-L$^\dag$~\cite{convnext} & 232M & 749G & 256 & 81  & \textbf{58.0} & \begin{tabular}{@{}c@{}} 64.2 \end{tabular} & \begin{tabular}{@{}c@{}} \textbf{48.6} \end{tabular} & --- & \gr{---} & --- \\
    
    \midrule
    \midrule
    \multicolumn{9}{l}{\textit{\textbf{Joint Training}}}\\
    \midrule
    \midrule

    \begin{tabular}{@{}l@{}}\textbf{\modelname} \end{tabular} & Swin-L$^\dag$~\cite{swin-T} & 219M & 891G & 150 & 100  & \begin{tabular}{@{}c@{}} \textbf{57.9} \end{tabular} & \begin{tabular}{@{}c@{}} \textbf{64.4} \end{tabular} & \begin{tabular}{@{}c@{}} 48.0  \end{tabular} & \begin{tabular}{@{}c@{}} \textbf{49.0} \end{tabular} & \gr{48.9} & \begin{tabular}{@{}c@{}} \textbf{67.4} \end{tabular} \\
    \midrule

    \begin{tabular}{@{}l@{}}\textbf{\modelname} \end{tabular} & DiNAT-L$^\dag$~\cite{dinat} & 223M & 736G & 150 & 100  & \begin{tabular}{@{}c@{}} \textbf{58.0} \end{tabular} & \begin{tabular}{@{}c@{}} \textbf{64.3} \end{tabular} & \begin{tabular}{@{}c@{}} \textbf{48.4}  \end{tabular} & \begin{tabular}{@{}c@{}} \textbf{49.2} \end{tabular} & \gr{49.2} & \begin{tabular}{@{}c@{}} \textbf{68.1} \end{tabular} \\
    
    \bottomrule
\end{tabular}}
  \vspaceundertab
  \caption{\textbf{SOTA Comparison on COCO val2017 set.} $^\dag$: Imagenet-22k pretrained; $^\ddag$: retrained model result; $^*$: trained with batch size 64. \modelname outperforms the individually trained Mask2Former~\cite{mask2former} on all metrics. We evaluate the AP score on instance ground truths derived from the panoptic annotations. Mask2Former's performance with 150 queries is not listed, as its performance degrades with 150 queries. We compute FLOPs using 100 validation COCO images (varying sizes). \gr{AP$^\text{instance}$ represents evaluation on the original instance annotations.} 
      }
    \label{tab:sota_coco}
\vspaceundercaption
\end{table*}

\subsection{Datasets and Evaluation Metrics}

\noindent
\textbf{Datasets.} We experiment on three widely used datasets that support all three: semantic, instance, and panoptic segmentation tasks. \textbf{Cityscapes}~\cite{cityscapes} consists of a total 19 (11 ``stuff" and 8 ``thing") classes with 2,975 training, 500 validation and 1,525 test images. \textbf{ADE20K}~\cite{ade20k} is another benchmark dataset with 150 (50 ``stuff" and 100 ``thing") classes among the 20,210 training and 2,000 validation images. \textbf{COCO}~\cite{coco} has 133 (53 ``stuff" and 80 ``thing") classes with 118k training and 5,000 validation images.

\noindent
\textbf{Evaluation Metrics.} For all three image segmentation tasks, we report the \textbf{PQ}~\cite{pq}, \textbf{AP}~\cite{coco}, and \textbf{mIoU}~\cite{everingham2015pascal} scores. Since we only have a single model for all three tasks, we use the value of the \texttt{task} token to decide the scores to consider. For \emph{e.g.}, when \texttt{task} is \texttt{panoptic}, we report the \textbf{PQ} score and similarly we report \textbf{AP} and \textbf{mIoU} scores when \texttt{task} is \texttt{instance} and \texttt{semantic}, respectively.

\subsection{Main Results}

\noindent
\textbf{ADE20K.} We compare \modelname with the existing state-of-the-art pseudo-universal and specialized architectures on the ADE20K~\cite{ade20k} val dataset in \cref{tab:sota_ade20k}. With the standard Swin-L$^\dag$ backbone, \modelname, while being trained only once, outperforms Mask2Former's~\cite{mask2former} individually trained models on all three image segmentation tasks and sets a new state-of-the-art performance when compared with other methods using the same backbone.

\noindent
\textbf{Cityscapes.} We compare \modelname with the existing state-of-the-art pseudo-universal and specialized architectures on the Cityscapes~\cite{ade20k} val dataset in \cref{tab:sota_city}. With Swin-L$^\dag$ backbone, \modelname outperforms Mask2Former with a $+\mathbf{0.6}\%$ and $+\mathbf{1.9}\%$ improvement on the \textbf{PQ} and \textbf{AP} metrics, respectively. Additionally, with ConvNeXt-L$^\dag$ and ConvNeXt-XL$^\dag$ backbone, \modelname sets a new state-of-the-art of $68.5\%$ PQ and $46.7\%$ AP, respectively.   

\noindent
\textbf{COCO.} We compare \modelname with the existing state-of-the-art pseudo-universal and specialized architectures on the COCO~\cite{coco} val2017 dataset in \cref{tab:sota_coco}. With Swin-L$^\dag$ backbone, \modelname performs on-par with the individually trained Mask2Former~\cite{mask2former} with a $+0.1\%$ improvement in the PQ score. Due to the discrepancies between the panoptic and instance annotations in COCO~\cite{coco}, we evaluate the AP score using the instance ground truths derived from the panoptic annotations. We provide more information in the appendix. Following ~\cite{mask2former}, we evaluate mIoU on semantic ground truths derived from panoptic annotations.

\subsection{Ablation Studies}
\label{subsec:ablations}

We analyze \modelname's components through a series of ablation studies. Unless stated otherwise, we ablate with Swin-L$^\dag$ \modelname on the Cityscapes~\cite{cityscapes} dataset.

% \smallskip
\noindent
\textbf{Task-Conditioned Architecture.} We validate the importance of the task token ($\mathbf{Q}_\text{task}$), initializing the queries with repetitions of the task token (task-guided query init.) and the learnable text context ($\mathbf{Q}_\text{ctx}$)  by removing each component one at a time in \cref{tab:ablat_component}. Without the task token, we observe a significant drop in the AP score ($-2.7\%$). Furthermore, using a learnable text context ($\mathbf{Q}_\text{ctx}$) leads to an improvement of $+4.5\%$ in the PQ score, proving its significance. Lastly, initializing the queries as repetitions of the task token (task-guided query init.) instead of using an all-zeros initialization~\cite{detr} leads to an improvement of $+1.4\%$ in the PQ and $+1.1\%$ in the AP score, indicating the importance of task-conditioning the initialization of the queries.

% \smallskip
\noindent
\textbf{Contrastive Query Loss.} We report results without the query-text contrastive loss ($\mathcal{L}_{\mathbf{Q} \leftrightarrow {\mathbf{Q}_\text{text}}}$) in \cref{tab:ablat_loss}. We observe that the contrastive loss significantly benefits the PQ ($+8.4\%$) and AP ($+3.2\%$) scores. We also conduct experiments substituting our query-text contrastive loss with a classification loss ($\mathcal{L}_\text{cls}$) on the queries. $\mathcal{L}_\text{cls}$ can be regarded as a straightforward alternative for $\mathcal{L}_{\mathbf{Q} \leftrightarrow {\mathbf{Q}_\text{text}}}$ as the both provide supervision for the number of masks for each class present in the image. However, we observe significant drops on all the metrics ($-0.8\%$ PQ, $-0.9\%$ AP and $-0.4\%$ mIoU) using the classification loss instead of the contrastive loss. We attribute the drops to the inability of the classification loss to capture the inter-task differences effectively.

\begin{table}[t!]\setlength{\tabcolsep}{10pt}
  \centering
  \resizebox{1.\linewidth}{!}{
  \begin{tabular}{l|ccc}

& PQ & AP & mIoU \\
\midrule

\textbf{\modelname} \dt{ours} & \textbf{67.2} & \textbf{45.6} & \textbf{83.0} \\
\midrule

$-$ task-token ($\mathbf{Q}_\text{task}$) & 66.5 \pp{\dt{-0.7}} & 43.3 \pp{\dt{-2.3}} & 82.9 \pp{\dt{-0.1}} \\

$-$ learnable text context ($\mathbf{Q}_\text{ctx}$) & 62.7 \pp{\dt{-4.5}} & 45.0 \pp{\dt{-0.6}} & 82.8 \pp{\dt{-0.2}}  \\

$-$ task-guided query init. & 65.8 \pp{\dt{-1.4}} & 44.5 \pp{\dt{-1.1}} & 83.1 \az{\dt{+0.1}} \\

\end{tabular}}
  \vspaceundertab
  \caption{\textbf{Ablation on Components.} A task-conditioned architecture significantly improves the AP scores and using learnable text context improves the PQ score.
      }
    \label{tab:ablat_component}
\vspaceundercaption
\end{table}

\begin{table}[t!]\setlength{\tabcolsep}{10pt}
  \centering
  \resizebox{1.\linewidth}{!}{
  \begin{tabular}{l | ccc|c}

& PQ & AP & mIoU. & \#param. \\
\midrule

\textbf{contrastive-loss} \dt{ours} & \textbf{67.2} & \textbf{45.6} & \textbf{83.0} & 219M \\
\midrule

query classification-loss & 66.4 \pp{\dt{-0.8}} & 44.7 \pp{\dt{-0.9}} & 82.6 \pp{\dt{-0.4}} & 219M \\

no contrastive-loss & 58.8 \pp{\dt{-8.4}} & 42.4 \pp{\dt{-3.2}} & 82.5 \pp{\dt{-0.5}} & 219M \\
\end{tabular}}
  \vspaceundertab
  \caption{\textbf{Ablation on Loss.} The contrastive loss is essential for learning the inter-task distinctions during training.
      }
    \label{tab:ablat_loss}
\vspaceundercaption
\end{table}

\begin{table}[t!]\setlength{\tabcolsep}{10pt}
  \centering
  \resizebox{1.\linewidth}{!}{
  \begin{tabular}{l|ccc}

& PQ & AP & mIoU \\
\midrule

\textbf{``a photo with a \{\texttt{CLS}\}"} \dt{ours} & \textbf{67.2} & \textbf{45.6} & \textbf{83.0} \\
\midrule

``a photo with a \{\texttt{CLS}\} \{\texttt{TYPE}\}"  & 65.4 \pp{\dt{-1.8}} & 44.5 \pp{\dt{-1.1}} & 82.8 \pp{\dt{-0.2}} \\

``\{\texttt{CLS}\}" & 66.6 \pp{\dt{-0.6}} & 44.7 \pp{\dt{-0.9}} & 82.5 \pp{\dt{-0.5}} \\

\end{tabular}}
  \vspaceundertab
  \caption{\textbf{Ablation on Input Text Templates.}  The template for the input text list entries is a critical factor for good performance. \texttt{CLS} represents the class name for the object and \texttt{TYPE} stands for the task-dependent object type.
      }
    \label{tab:ablat_template}
\vspaceundercaption
\end{table}

% \smallskip
\noindent
\textbf{Input Text Template.} We study the importance of the template choice for the entries in the text list ($\mathbf{T}_\text{list}$) in \cref{tab:ablat_template}. We experiment with ``a photo with a \{\texttt{CLS}\} \{\texttt{TYPE}\}" template for our text entries where \texttt{CLS} is the class name for the object mask and \texttt{TYPE} is the task-dependent class-type: ``stuff" for amorphous masks (panoptic and semantic task) and ``thing" for all distinct object masks. We also experiment with the identity template ``\{\texttt{CLS}\}". Our choice of the template: ``a photo with a \{\texttt{CLS}\}" gives a strong performance as a baseline. We believe more exploration in the text template space could help in improving the performance further.

% \smallskip
\noindent
\textbf{Task Conditioned Joint Training.} As a baseline for comparison, we train a Swin-L$^\dag$ Mask2Former-Joint with our joint training strategy, \emph{i.e.}, uniformly sampling each task's GT on the ADE20K~\cite{ade20k} dataset. We compare the Mask2Former-Joint baseline with our Swin-L$^\dag$ \modelname in \cref{tab:ablat_joint}. We train both models for 160k iterations with a batch size of 16. Our \modelname achieves a $+1.1\%$, $+2.2\%$, and $+0.8\%$ improvement on the PQ, AP and mIoU metrics, respectively, proving the importance of our architecture design for practical multi-task joint training.

\begin{table}[t!]\setlength{\tabcolsep}{10pt}
  \centering
  \resizebox{1.\linewidth}{!}{
  \begin{tabular}{l |ccc|c}

& PQ & AP & mIoU & \#param. \\
\midrule

\textbf{\modelname} \dt{ours} & \textbf{49.8} & \textbf{35.9}  & \textbf{57.0} & 219M \\
\midrule

Mask2Former-Joint & 48.7 \pp{\dt{-1.1}} & 33.7 \pp{\dt{-2.2}} & 56.2 \pp{\dt{-0.8}} & 216M\\

% Mask2Former-Panoptic & 53.8 \pp{\dt{-0.6}} & 43.3 \pp{\dt{-1.9}} & 63.6 \pp{\dt{-0.4}} & 47M\\

\end{tabular}}
  \vspaceundertab
  \caption{\textbf{Ablation on Joint Training.} Our \modelname significantly beats the baseline's AP, PQ and mIoU scores. We report results with Swin-L$^\dag$~\cite{swin-T} backbone trained for 160k iterations on the ADE20K~\cite{ade20k} dataset. 
      }
    \label{tab:ablat_joint}
\vspaceundercaption
\end{table}

\begin{table}[t!]\setlength{\tabcolsep}{10pt}
  \centering
  \resizebox{1.\linewidth}{!}{
  \begin{tabular}{l|ccc|cc}

 Task Token Input & PQ & PQ$^\text{Th}$ & PQ$^\text{St}$ & AP & mIoU \\
\midrule

the task is \texttt{panoptic} & 49.3 & 49.6 & 50.2 & \mg{35.8}  & \mg{57.0} \\
the task is \texttt{instance} & \mg{33.1} & \mg{48.8} & \mg{1.5} & 35.9 & \mg{26.4} \\
the task is \texttt{semantic} & \mg{40.4} & \mg{35.5} & \mg{50.2} & \mg{25.3} & 57.0 \\

\end{tabular}}
  \vspaceundertab
  \caption{\textbf{Ablation on Task Token Input}. Our \modelname is sensitive to the input task token value. We report results with Swin-L$^\dag$ \modelname on the ADE20K~\cite{ade20k} val set. The numbers in \mg{pink} denote results on secondary \texttt{task} metrics.
      }
    \label{tab:ablat_task}
\vspaceundercaption
\end{table}

\begin{figure}[t!]
  \centering
\includegraphics[width=\linewidth]{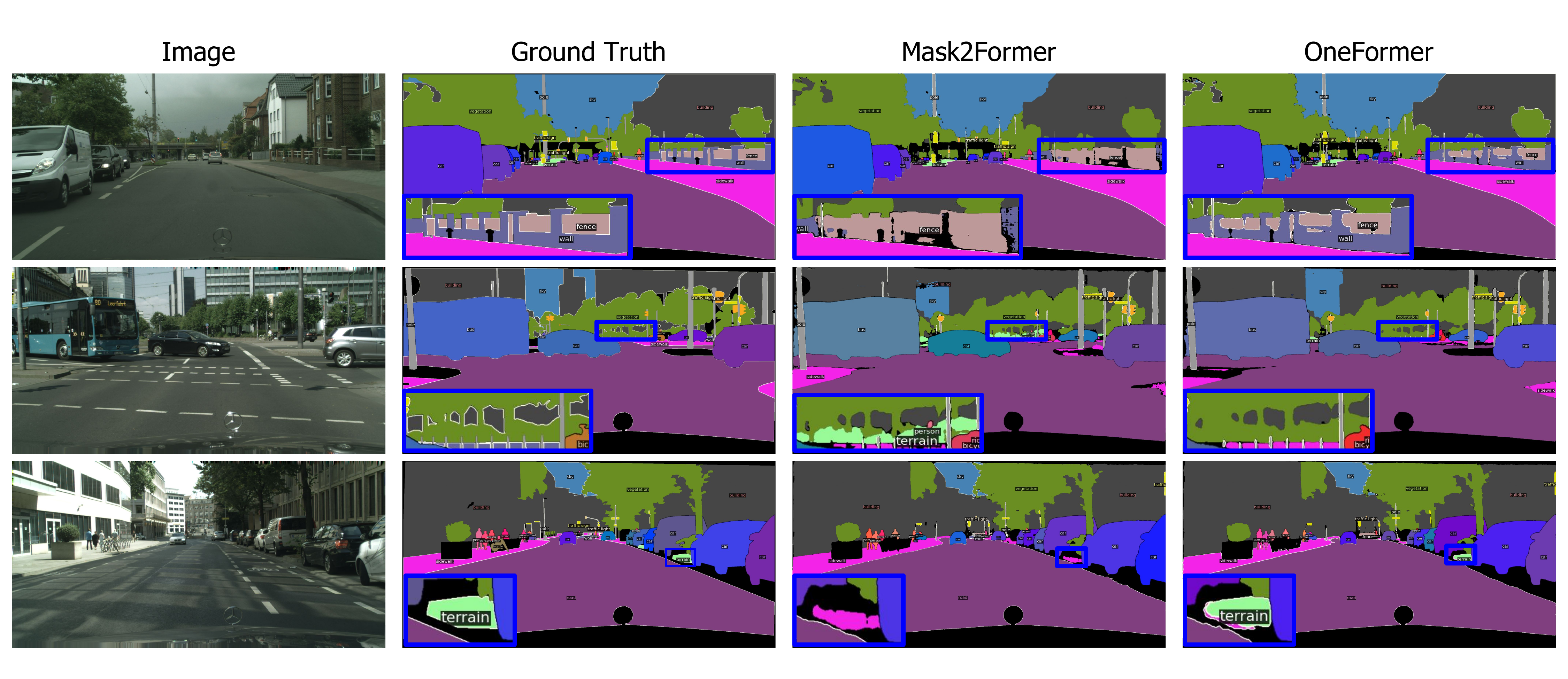}
  \vspaceunderfigext
  \caption{
      \textbf{Reduced Category Misclassifications.} Our \modelname segments the regions (inside \bl{blue} boxes) with similar classes more accurately than Mask2Former~\cite{mask2former}. \textbf{\pp{Zoom in for best view.}}
      }
  \label{fig:class_fig}
\vspaceundercaption
\end{figure}

% \smallskip
\noindent
\textbf{Task Token Input.} We demonstrate that our framework is sensitive to the task token input by setting the value of \{\texttt{task}\} during inference as panoptic, instance, or semantic in \cref{tab:ablat_task}. We report results with our Swin-L$^\dag$ \modelname trained on ADE20K~\cite{ade20k} dataset. We observe a significant drop in the PQ and mIoU metrics when \texttt{task} is instance compared to panoptic. Moreover, the PQ$^\text{St}$ drops to $1.5\%$, and there is only a $-0.8\%$ drop on PQ$^\text{Th}$ metric, proving that the network learns to focus majorly on the distinct ``thing" instances when the \texttt{task} is instance. Similarly, there is a sizable drop in the PQ, PQ$^\text{Th}$ and AP metrics for the semantic task with PQ$^\text{St}$ staying the same, showing that our framework can segment out amorphous masks for ``stuff" regions but does not predict different masks for ``thing" objects. Therefore, OneFormer dynamically learns the inter-task distinctions which is critical for a train-once multi-task architecture. We include qualitative analysis on the task dynamic nature of \modelname in the appendix.

% \smallskip
\noindent
\textbf{Reduced Category Misclassifications.} Our query-text contrastive loss helps OneFormer learn the inter-task distinctions and reduce the number of category misclassifications in the predictions. Mask2Former incorrectly predicts ``wall" as ``fence" in the first row, ``vegetation" as ``terrain", and ``terrain" as ``sidewalk". At the same time, our \modelname produces more accurate predictions in regions (inside \bl{blue} boxes) with similar classes, as shown in \cref{fig:class_fig}.
\section{Conclusion}

In this work, we present \modelname, a new multi-task universal image segmentation framework with transformers and task-guided queries to unify semantic, instance, and panoptic segmentation with a single universal architecture, a single model, and training on a single dataset. Our jointly trained single \modelname model outperforms the individually trained specialized Mask2Former models, the previous single-architecture state of the art, on all three segmentation tasks across major datasets. Consequently, \modelname can cut training time, weight storage, and inference hosting requirements down to a third, making image segmentation more accessible. 
We believe \modelname is a significant step towards making
image segmentation more universal and accessible and will support further research in this direction by open-sourcing our codes and models.

% We train our \modelname only once using our proposed task-conditioned joint training strategy.

% More specifically, we use a text task token input and a query-text contrastive loss on the queries to condition \modelname on the segmentation task (panoptic, instance, or semantic).

% \noindent
% \textbf{Limitations}: 

\vspace{-0.4cm}
\paragraph{Acknowledgments.} We thank Intelligence Advanced Research Projects Activity (IARPA), University of Oregon, University of Illinois at Urbana-Champaign, and Picsart AI Research (PAIR) for their generous support that made this work possible.

%%%%%%%%% REFERENCES
\bibliographystyle{ieee_fullname}
\bibliography{main}

\newpage
\appendix
\begin{center}{\bf \Large Appendix}\end{center}\vspace{-2mm}
\renewcommand{\thetable}{\Roman{table}}
\renewcommand{\thefigure}{\Roman{figure}}
\setcounter{table}{0}
\setcounter{figure}{0}

\Crefname{appendix}{Appendix}{Appendixes}

\section{Implementation Details}

We implement our framework using the Detectron2~\cite{wu2019detectron2} library.

\noindent
\textbf{Multi-Scale Feature Modeling.} We adopt the settings from \cite{mask2former} for modeling the image pixel-level features. More specifically, we use 6 MSDeformAttn~\cite{deformable-detr} inside our pixel decoder, applied to feature maps with resolutions $1/8$, $1/16$, and $1/32$ of the original image. We use lateral connections and upsampling to aggregate the multi-scale features to a final $1/4$ resolution scale. We map all the features to a hidden dimension of $256$.

\noindent
\textbf{Unified Task-Conditioned Query Formulation.} We initialize the $N-1$ queries as repetitions of task-token, $\mathbf{Q}_\text{task}$. Unless stated otherwise, we set $N=250$ and $N_\text{ctx}=16$. Our text tokenizer and text encoder are the same as \cite{groupvit}. We use a single linear layer to project the tokenized task input, followed by a layer-norm to obtain $\mathbf{Q}_\text{task}$.

\noindent
\textbf{Task-Dynamic Mask and Class Prediction Formation.} Following ~\cite{mask2former}, we set $L=3$ inside the transformer decoder. Therefore, we have a total of $3L$ (9) stages inside our transformer decoder. We also calculate an auxiliary loss on each intermediate class and mask predictions after every transformer decoder stage~\cite{mask2former}.

\noindent
\textbf{Training Settings.} We train our model with a batch size of 16. When training on ADE20K~\cite{ade20k} and Cityscapes~\cite{cityscapes}, we use the AdamW~\cite{adamW} optimizer with a base learning rate of $0.0001$, poly learning rate decay and weight decay $0.1$. We use a crop size of $512\!\times\!512$ and $512\!\times\!1024$ on ADE20K and Cityscapes, respectively. We train for 90k and 160k iterations on Cityscapes and ADE20K, respectively. For data augmentation, we use shortest edge resizing, fixed size cropping, and color jittering followed by a random horizontal flip.

When training on COCO~\cite{coco}, we use a step learning rate schedule along with the AdamW~\cite{adamW} optimizer, a base learning rate of $0.0001$, $10$ warmup iterations, and a weight decay of $0.05$. We decay the learning rate at $0.9$ and $0.95$ fractions of the total number of training steps by a factor of $10$. We train for a total of 100 epochs with LSJ augmentation~\cite{simple-copy-paste, simple-training} with a random scale sampled from the range $0.1$ to
$2.0$ followed by a fixed size crop to $1024\!\times\!1024$ resolution.

\noindent
\textbf{Evaluation Settings.} We follow the same evaluation settings as Mask2Former~\cite{mask2former}. Unless stated otherwise, we report results for the single-scale inference setting. Unlike the training stage, during evaluation, we use the ground-truth annotations from the respective \texttt{task} GT labels to calculate the metric scores instead of deriving the labels from the panoptic annotations. Additionally, we set the value of \texttt{task} in ``the task is \{\texttt{task}\}" as panoptic, instance and semantic to obtain the corresponding task predictions.

\section{Additional Ablations}
\noindent
\textbf{Ablation on Number of Queries.} We study the effect of the different number of queries on the COCO dataset in \cref{tab:ablat_q}. We conduct experiments using the ResNet-50 (R50)~\cite{resnet} backbone and train for 50 epochs. We find that $N=150$ performs the best. 

Additionally, we tune the number of queries on the Swin-L$^\dag$ backbone separately. During our experiments, we found that $N=250$ is the best setting with Swin-l$^\dag$ on ADE20K~\cite{ade20k} and Cityscapes~\cite{cityscapes} datasets. On COCO~\cite{coco}, $N=150$ gives the best performance with Swin-L$^\dag$. We also noticed that with smaller backbones like R50~\cite{resnet}, $N=150$ is the optimal setting on the ADE20K~\cite{ade20k} dataset.

\begin{table}[t!]\setlength{\tabcolsep}{10pt}
  \centering
  \resizebox{0.9\linewidth}{!}{
  \begin{tabular}{c|ccc|c}
\#queries & PQ & AP & mIoU & \#param. \\
\midrule
100 & 51.3 & 41.9 & 60.8 & 47M\\
120 & 51.0 & 42.0 & 60.8 & 47M \\
150 & \textbf{51.5} & \textbf{42.5} & \textbf{61.2} & 47M \\
200 & 51.3 & \textbf{42.5} & 60.0 & 47M \\
\end{tabular}}
  \vspaceundertab
  \caption{\textbf{Ablation on Number of Queries.} We find $N=150$ performs best on the COCO dataset.
      }
    \label{tab:ablat_q}
    \vspaceundercaption
\end{table}

\begin{table}[!t]\setlength{\tabcolsep}{10pt}
  \centering
  \resizebox{0.9\linewidth}{!}{
  \begin{tabular}{c|ccc|c}
$N_\text{ctx}$ & PQ & AP & mIoU & \#param. \\
\midrule
0 & 41.7 & \textbf{27.5} & 46.5 & 47M\\
8 & 41.0 & 27.2 & 46.5 & 47M \\
16 & \textbf{41.9} & 27.3 & \textbf{47.3} & 47M \\
32 & 41.7 & \textbf{27.5} & 46.8 & 47M \\
\end{tabular}}
  \vspaceundertab
  \caption{\textbf{Ablation on number of learnable text context embeddings.} We find $N_\text{ctx}=16$ performs best.
      }
    \label{tab:ablat_q_ctx}
    \vspaceundercaption
\end{table}

\begin{table}[t!]\setlength{\tabcolsep}{10pt}
  \centering
  \resizebox{0.9\linewidth}{!}{
  \begin{tabular}{c|cccc}
contrastive-loss weight & PQ & AP & mIoU \\
\midrule
$\lambda_{\mathbf{Q} \leftrightarrow {\mathbf{Q}_\text{text}}}=0.0$ & 51.1 & 42.1 & 60.2 \\
$\lambda_{\mathbf{Q} \leftrightarrow {\mathbf{Q}_\text{text}}}=0.5$ & \textbf{51.5} & \textbf{42.5} & \textbf{61.2} \\
$\lambda_{\mathbf{Q} \leftrightarrow {\mathbf{Q}_\text{text}}}=1.0$ & 50.7 & 42.0 & 60.5 \\

\end{tabular}}
  \vspaceundertab
  \caption{\textbf{Ablation on Contrastive Loss' Weight.} We find $\lambda_{\mathbf{Q} \leftrightarrow {\mathbf{Q}_\text{text}}}=0.5$ gives the best performance.
      }
    \label{tab:ablat_contrastive_weight}
    \vspace{-0.3cm}
\end{table}

\begin{figure*}[t!]
  \centering
\includegraphics[width=\linewidth]{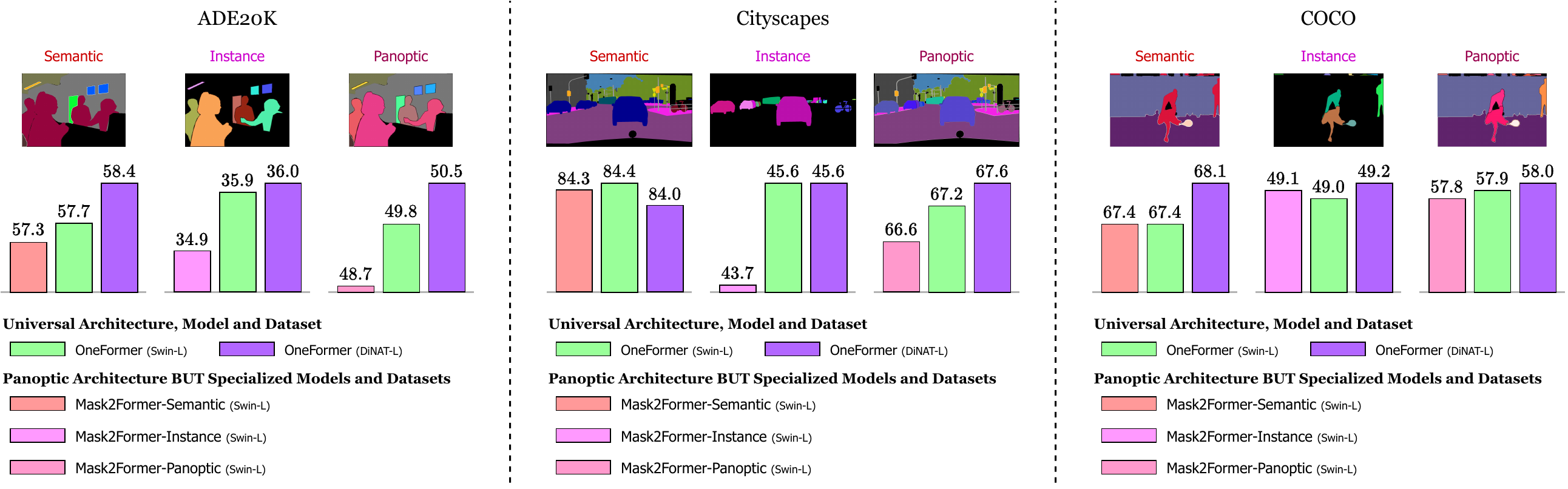}
  \vspaceunderfigext
  \caption{
      \textbf{Comparison to Swin-L Mask2Former~\cite{mask2former} across leaderboards.} Our single \modelname model outperforms Mask2Former~\cite{mask2former}, the previous single architecture SOTA system on ADE20K val~\cite{ade20k}, Cityscapes val~\cite{cityscapes}, and COCO val2017~\cite{coco} for all three segmentation tasks. With DiNAT-L \modelname, we achieve even more improvements.}
  \label{fig:plots}
\vspaceundercaption
\end{figure*}
\noindent
\textbf{Ablation on Contrastive Loss' Weight.} We run ablations on the weight for the contrastive loss' weight on the COCO dataset in \cref{tab:ablat_contrastive_weight}. We conduct our experiments using the ResNet-50 (R50)~\cite{resnet} backbone and train for 50 epochs. We find that $\lambda_{\mathbf{Q} \leftrightarrow {\mathbf{Q}_\text{text}}}=0.5$ is the optimal weight setting.

\noindent
\textbf{Ablation on Number of Learnable Text Context Embeddings.} We study the effect of different number of learnable text context embeddings on the ADE20K~\cite{ade20k} dataset in \cref{tab:ablat_q_ctx}. We conduct our experiments using the ResNet-50 (R50)~\cite{resnet} backbone and train for 160k iterations. We find that $N_\text{ctx}=16$ performs best.

% \textit{Note that Mask2Former's performance with 120 queries is not listed as its performance will degrade when using 120 queries}.

\section{Individual Training}

In this section, we analyze our \modelname's performance with individual training on the panoptic, instance, and semantic segmentation task. For this study, we conduct experiments with the ResNet-50 (R50)~\cite{resnet} backbone on the ADE20K~\cite{ade20k} dataset. We train all models for 160k iterations with a batch size of 16.

As shown in \cref{tab:supp_ade20k}, \modelname outperforms Mask2Former~\cite{mask2former} (the previous SOTA pseudo-universal image segmentation method) with every training strategy. Furthermore, with joint training, Mask2Former~\cite{mask2former} suffers a significant drop in performance, and \modelname achieves the highest PQ, AP and mIoU scores.

In order to train \modelname on a single task, we set the value of \texttt{task} as that of the corresponding task in our task token input: ``the task is \{\texttt{task}\}" for the samples during training. Therefore, under Panoptic Training, only panoptic ground truth labels will be used, and similarly, for Semantic and Instance Training, only semantic and instance ground truth labels shall be used, respectively. The joint training strategy remains the same as described in Sec \pp{3.1} (main text) with uniform sampling for each task-specific ground truth label. Note that for training \modelname, we derive all ground truth labels from the panoptic annotations.

\section{Analysis on the Task-Dynamic Nature of \modelname}

We analyze \modelname's ability to capture the inter-task differences by changing the value of \{\texttt{task}\} in the task token input: ``the task is \{\texttt{task}\}" as panoptic, instance, or semantic, during inference. We report quantitative report results with our Swin-L$^\dag$ \modelname trained on Cityscapes~\cite{cityscapes} dataset in \cref{tab:supp_task_city}. When we set \texttt{task} as ``instance", we observe that PQ$^\text{St}$ drops to $0.0\%$, and there is only a $-0.2\%$ drop on PQ$^\text{Th}$ metric as compared to the setting when \texttt{task} is panoptic. This observation proves that \modelname learns to change its feed-forward output depending on the task dynamically. Similarly, there is a sizable drop in the PQ, PQ$^\text{Th}$ and AP metrics for the semantic task with PQ$^\text{St}$ improving by $+0.2\%$ showing that our framework can segment out amorphous masks for ``stuff" regions but does not predict different masks for ``thing" objects.

\begin{table*}[!t]\setlength{\tabcolsep}{10pt}
  \centering
  \resizebox{0.7\linewidth}{!}{
  \begin{tabular}{l|l|ccc}
   training strategy & method & PQ & AP & mIoU \\
    \midrule
    
     \multirow{ 2}{*}{Panoptic Training} & \begin{tabular}{@{}l@{}} Mask2Former~\cite{mask2former} \end{tabular}  & 40.7 & 25.2 & 45.6 \\
    
     & \begin{tabular}{@{}l@{}}\textbf{\modelname} \small{(ours)} \end{tabular} & \textbf{41.4} \az{\dt{+0.7}} & \textbf{27.0} \az{\dt{+1.8}} &\textbf{46.1} \az{\dt{+0.5}} \\
     \midrule
    
    \multirow{ 2}{*}{Instance Training} & \begin{tabular}{@{}l@{}} Mask2Former~\cite{mask2former} \end{tabular} & --- & 26.4 & --- \\
    
     & \begin{tabular}{@{}l@{}}\textbf{\modelname} \small{(ours)} \end{tabular} & --- & \textbf{26.7} \az{\dt{+0.3}} & --- \\
     \midrule
     
     \multirow{ 2}{*}{Semantic Training} & \begin{tabular}{@{}l@{}} Mask2Former~\cite{mask2former} \end{tabular} & --- & --- & 47.2 \\
    
     & \begin{tabular}{@{}l@{}}\textbf{\modelname} \small{(ours)} \end{tabular} & --- & --- & \textbf{47.3} \az{\dt{+0.1}} \\
     \midrule
    
    \multirow{ 2}{*}{Joint Training} & \begin{tabular}{@{}l@{}} Mask2Former$^\dag$~\cite{mask2former} \end{tabular} & 40.8 & 25.7 & 46.6 \\
    
     & \begin{tabular}{@{}l@{}}\textbf{\modelname} \small{(ours)} \end{tabular} & \textbf{41.9} \az{\dt{+1.1}} & \textbf{27.3} \az{\dt{+1.6}} &\textbf{47.3} \az{\dt{+0.7}} \\
   
    \bottomrule
    
\end{tabular}}
  \vspaceundertab
  \caption{\textbf{Comparison between Individual and Joint Training.} Unlike Mask2Former~\cite{mask2former} which shows large variance in performance among the different training strategies, \modelname performs fairly well under all training strategies and outperforms Mask2Former~\cite{mask2former}. We train all models with R50~\cite{resnet} backbone on the ADE20K~\cite{ade20k} dataset for 160k iterations.  $^\dag$ We retrain our own Mask2Former~\cite{mask2former} using the joint training strategy.
      }
    \label{tab:supp_ade20k}
    \vspaceundercaption
\end{table*}

\begin{table}[t!]\setlength{\tabcolsep}{10pt}
\centering
\resizebox{1.\linewidth}{!}{
\begin{tabular}{l|ccc|cc}

 Task Token Input & PQ & PQ$^\text{Th}$ & PQ$^\text{St}$ & AP & mIoU \\
\midrule

the task is \texttt{panoptic} & 67.2 & 61.0 & 71.7 & \mg{45.3}  & \mg{83.0} \\
the task is \texttt{instance} & \mg{25.6} & \mg{60.8} & \mg{0.0} & 45.6 & \mg{6.3} \\
the task is \texttt{semantic} & \mg{56.9} & \mg{36.2} & \mg{71.9} & \mg{27.2} & 83.0 \\

\end{tabular}}
\vspaceundertab
\caption{\textbf{Quantitative Analysis on Task Dynamic Nature of \modelname}. Our \modelname is sensitive to the input task token value. We report results with Swin-L$^\dag$ \modelname on the Cityscapes~\cite{cityscapes} val set. The numbers in \mg{pink} denote results on secondary \texttt{task} metrics.
  }
\label{tab:supp_task_city}
\vspaceundercaption
\end{table}

\begin{table*}[t!]\setlength{\tabcolsep}{10pt}
  \centering
  \resizebox{1.\linewidth}{!}{
  \begin{tabular}{ll|ccc|cc|cc}
   \textbf{Method} & \textbf{Backbone} & \#\textbf{Params} & \textbf{Crop Size} & \begin{tabular}{@{}c@{}} \textbf{Extra Data} \end{tabular} & \textbf{PQ} & \textbf{AP} & \begin{tabular}{@{}c@{}} \textbf{mIoU} \\ (s.s.) \end{tabular} & \begin{tabular}{@{}c@{}} \textbf{mIoU} \\ (m.s.) \end{tabular} \\

    \midrule
    \midrule
    \multicolumn{9}{l}{\textit{\textbf{Individual Training}}}\\
    \midrule
    \midrule

    \begin{tabular}{@{}l@{}}Mask2Former~\cite{mask2former} \end{tabular} & BEiT-3~\cite{beit-3} & 1.9B & $896\!\times\!896$ & \cmark & --- & \begin{tabular}{@{}c@{}} --- \end{tabular} & \begin{tabular}{@{}c@{}} 62.0 \end{tabular} & 62.8 \\

    \begin{tabular}{@{}l@{}}UPerNet~\cite{upernet} \end{tabular} & FD-SwinV2-G~\cite{wei2022FD} & $>$3B & $896\!\times\!896$ & \cmark & \begin{tabular}{@{}c@{}} --- \end{tabular} & \begin{tabular}{@{}c@{}} --- \end{tabular} & \begin{tabular}{@{}c@{}} --- \end{tabular} & 61.4 \\

    \begin{tabular}{@{}l@{}}Mask DINO~\cite{li2022mask-dino} \end{tabular} & Swin-L~\cite{swin-T} & 223M & $896\!\times\!896$ & \cmark & \begin{tabular}{@{}c@{}} --- \end{tabular} & \begin{tabular}{@{}c@{}} --- \end{tabular} & \begin{tabular}{@{}c@{}} 59.5 \end{tabular} & 60.8 \\

     \begin{tabular}{@{}l@{}}{Mask2Former}~\cite{mask2former} \end{tabular} & {ViT-Adapter-L}~\cite{chen2022vitadapter} & {568M} & $896\!\times\!896$ & \cmark & \begin{tabular}{@{}c@{}} --- \end{tabular} & \begin{tabular}{@{}c@{}} --- \end{tabular} & \begin{tabular}{@{}c@{}} 59.4 \end{tabular} & 60.5 \\

     \begin{tabular}{@{}l@{}}UPerNet~\cite{upernet} \end{tabular} & SwinV2-G~\cite{swinv2} & $>$3B & $896\!\times\!896$ & \cmark & \begin{tabular}{@{}c@{}} --- \end{tabular} & \begin{tabular}{@{}c@{}} --- \end{tabular} & \begin{tabular}{@{}c@{}} 59.3 \end{tabular} & 59.9 \\

     \midrule
     \begin{tabular}{@{}l@{}}{UPerNet}~\cite{upernet} \end{tabular} & {ViT-Adapter-L}~\cite{chen2022vitadapter} & {571M} & $640\!\times\!640$ & \xmark & \begin{tabular}{@{}c@{}} --- \end{tabular} & \begin{tabular}{@{}c@{}} --- \end{tabular} & \begin{tabular}{@{}c@{}} 58.0 \end{tabular} & 58.4 \\
    
    \begin{tabular}{@{}l@{}}MSFaPN-Mask2Former~\cite{semask} \end{tabular} & SeMask Swin-L$^{\dag}$~\cite{semask} & --- & $640\!\times\!640$ & \xmark & \begin{tabular}{@{}c@{}} --- \end{tabular} & \begin{tabular}{@{}c@{}} --- \end{tabular} & \begin{tabular}{@{}c@{}} 57.0 \end{tabular} & 58.2 \\

    \begin{tabular}{@{}l@{}}FaPN-Mask2Former~\cite{fapn} \end{tabular} & Swin-L~\cite{swin-T} & --- & $640\!\times\!640$  & \begin{tabular}{@{}c@{}} \xmark \end{tabular} & \begin{tabular}{@{}c@{}} --- \end{tabular} & \begin{tabular}{@{}c@{}} --- \end{tabular} & \begin{tabular}{@{}c@{}} 56.4 \end{tabular} & \begin{tabular}{@{}c@{}} 57.7 \end{tabular} \\

     \begin{tabular}{@{}l@{}}SeMask Mask2Former~\cite{semask} \end{tabular} & SeMask Swin-L$^{\dag}$~\cite{semask} & --- & $640\!\times\!640$ & \xmark & \begin{tabular}{@{}c@{}} --- \end{tabular} & \begin{tabular}{@{}c@{}} --- \end{tabular} & \begin{tabular}{@{}c@{}} 56.4 \end{tabular} & 57.5 \\

    \begin{tabular}{@{}l@{}}Mask2Former-Semantic~\cite{fapn} \end{tabular} & Swin-L~\cite{swin-T} & 216M & $640\!\times\!640$  & \begin{tabular}{@{}c@{}} \xmark \end{tabular} & \begin{tabular}{@{}c@{}} --- \end{tabular} & \begin{tabular}{@{}c@{}} --- \end{tabular} & \begin{tabular}{@{}c@{}} 56.1 \end{tabular} & \begin{tabular}{@{}c@{}} 57.3 \end{tabular} \\

    \midrule
    \begin{tabular}{@{}l@{}}Mask2Former-Panoptic~\cite{mask2former} \end{tabular} & Swin-L~\cite{swin-T} & 216M & $640\!\times\!640$ & \xmark & \begin{tabular}{@{}c@{}} 48.1 \end{tabular} & \begin{tabular}{@{}c@{}} 34.2 \end{tabular} & \begin{tabular}{@{}c@{}} 54.5 \end{tabular} & \begin{tabular}{@{}c@{}} --- \end{tabular} \\

    \begin{tabular}{@{}l@{}}kMaX-DeepLab~\cite{kmax_deeplab} \end{tabular} & ConvNeXt-L$^\dag$~\cite{convnext} & 232M & $641\!\times\!641$ & \xmark & \begin{tabular}{@{}c@{}} 48.7 \end{tabular} & \begin{tabular}{@{}c@{}} --- \end{tabular} & \begin{tabular}{@{}c@{}} 54.8 \end{tabular} & --- \\

    \midrule
    \begin{tabular}{@{}l@{}}Mask2Former-Instance~\cite{mask2former} \end{tabular} & Swin-L~\cite{swin-T} & 216M & $640\!\times\!640$ & \xmark  & \begin{tabular}{@{}c@{}} --- \end{tabular} & \begin{tabular}{@{}c@{}} 34.9 \end{tabular} & \begin{tabular}{@{}c@{}} --- \end{tabular} & \begin{tabular}{@{}c@{}} --- \end{tabular} \\

    \begin{tabular}{@{}l@{}}kMaX-DeepLab~\cite{kmax_deeplab} \end{tabular} & ConvNeXt-L$^\dag$~\cite{convnext} & 232M & $1281\!\times\!1281$ & \xmark & \begin{tabular}{@{}c@{}} 50.9 \end{tabular} & \begin{tabular}{@{}c@{}} --- \end{tabular} & \begin{tabular}{@{}c@{}} 55.2 \end{tabular} & --- \\
    
    \midrule
    \midrule
    \multicolumn{9}{l}{\textit{\textbf{Joint Training}}}\\
    \midrule
    \midrule
    
    \begin{tabular}{@{}l@{}}\textbf{\modelname} \end{tabular} & Swin-L~\cite{swin-T} & 219M & $640\!\times\!640$ & \xmark  & \begin{tabular}{@{}c@{}} \textbf{49.8} \end{tabular} & \begin{tabular}{@{}c@{}} \textbf{35.9} \end{tabular} & \begin{tabular}{@{}c@{}} \textbf{57.0} \end{tabular} & \begin{tabular}{@{}c@{}} \textbf{57.7} \end{tabular} \\
    
    \begin{tabular}{@{}l@{}}\textbf{\modelname} \end{tabular} & Swin-L~\cite{swin-T} & 219M & $896\!\times\!896$ & \xmark  & \begin{tabular}{@{}c@{}} \textbf{51.1} \end{tabular} & \begin{tabular}{@{}c@{}} \textbf{37.6} \end{tabular} & \begin{tabular}{@{}c@{}} \textbf{57.4} \end{tabular} & \begin{tabular}{@{}c@{}} \textbf{58.3} \end{tabular} \\

    \begin{tabular}{@{}l@{}}\textbf{\modelname} \end{tabular} & Swin-L~\cite{swin-T} & 219M & $1280\!\times\!1280$ & \xmark  & \begin{tabular}{@{}c@{}} \textbf{51.4} \end{tabular} & \begin{tabular}{@{}c@{}} \textbf{37.8} \end{tabular} & \begin{tabular}{@{}c@{}} \textbf{57.0} \end{tabular} & \begin{tabular}{@{}c@{}} \textbf{57.7} \end{tabular} \\
    
    \midrule
    \begin{tabular}{@{}l@{}}\textbf{\modelname} \end{tabular} & ConvNeXt-L~\cite{convnext} & 220M & $640\!\times\!640$ & \xmark  & \begin{tabular}{@{}c@{}} \textbf{50.0} \end{tabular} & \begin{tabular}{@{}c@{}} \textbf{36.2} \end{tabular} & \begin{tabular}{@{}c@{}} \textbf{56.6} \end{tabular} & \begin{tabular}{@{}c@{}} \textbf{57.4} \end{tabular} \\
    
    \begin{tabular}{@{}l@{}}\textbf{\modelname} \end{tabular} & ConvNeXt-XL~\cite{convnext} & 372M & $640\!\times\!640$ & \xmark  & \begin{tabular}{@{}c@{}} \textbf{50.1} \end{tabular} & \begin{tabular}{@{}c@{}} \textbf{36.3} \end{tabular} & \begin{tabular}{@{}c@{}} \textbf{57.4} \end{tabular} & \begin{tabular}{@{}c@{}} \textbf{58.8} \end{tabular} \\
    
    \midrule
    \begin{tabular}{@{}l@{}}\textbf{\modelname} \end{tabular} & DiNAT-L~\cite{dinat} & 223M & $640\!\times\!640$ & \xmark  & \begin{tabular}{@{}c@{}} \textbf{50.5} \end{tabular} & \begin{tabular}{@{}c@{}} \textbf{36.0} \end{tabular} & \begin{tabular}{@{}c@{}} \textbf{58.3} \end{tabular} & \begin{tabular}{@{}c@{}} \textbf{58.4} \end{tabular} \\
    
    \begin{tabular}{@{}l@{}}\textbf{\modelname} \end{tabular} & DiNAT-L~\cite{dinat} & 223M & $896\!\times\!896$ & \xmark  & \begin{tabular}{@{}c@{}} \textbf{51.2} \end{tabular} & \begin{tabular}{@{}c@{}} \textbf{36.8} \end{tabular} & \begin{tabular}{@{}c@{}} \textbf{58.1} \end{tabular} & \begin{tabular}{@{}c@{}} \textbf{58.6} \end{tabular} \\

    \begin{tabular}{@{}l@{}}\textbf{\modelname} \end{tabular} & DiNAT-L~\cite{dinat} & 223M & $1280\!\times\!1280$ & \xmark  & \begin{tabular}{@{}c@{}} \textbf{51.5} \end{tabular} & \begin{tabular}{@{}c@{}} \textbf{37.1} \end{tabular} & \begin{tabular}{@{}c@{}} \textbf{58.2} \end{tabular} & \begin{tabular}{@{}c@{}} \textbf{58.7} \end{tabular} \\
    
    \bottomrule
    
\end{tabular}}
  \vspaceundertab
  \caption{\textbf{Comparison to methods on PwC Leaderboard on ADE20K val~\cite{ade20k}.} \modelname achieves new-state-of-the-art performances on all three segmentation tasks when compared with methods \textbf{not using extra training data}.
      }
    \label{tab:pwc_ade20k}
    \vspaceundercaption
\end{table*}

\begin{figure*}[t!]
\centering
\includegraphics[width=1.0\linewidth]{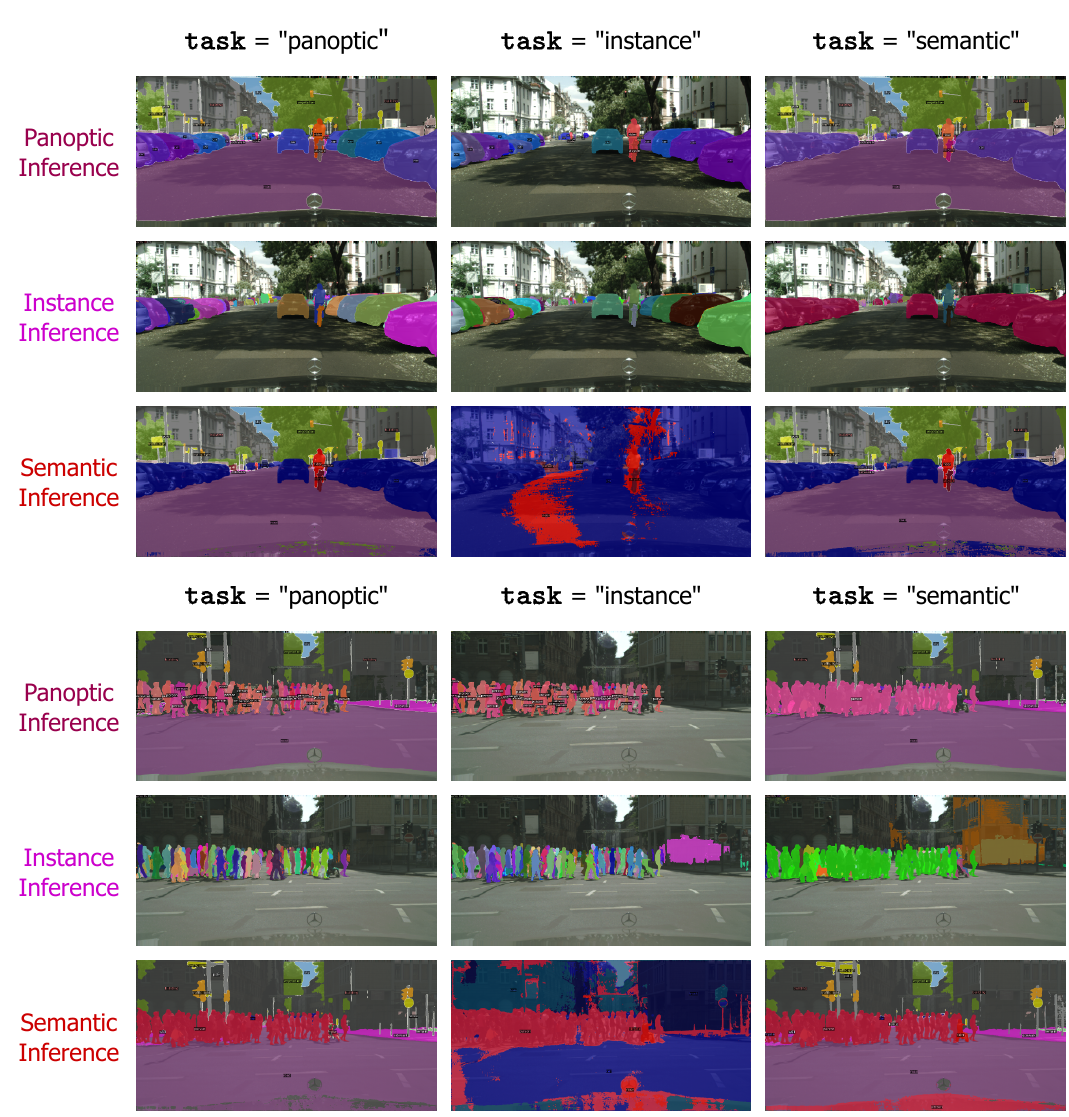}
\caption{
  \textbf{Qualitative Analysis on Task Dynamic Nature of \modelname.} When \texttt{task} = ``instance", the semantic inference outputs display fair detection of ``thing" regions and misclassifications for the ``stuff" regions. Similarly, when \texttt{task} = ``semantic", the distinct object masks are grouped into a single amorphous mask, as expected by the formulation of the semantic segmentation task. \textbf{\pp{Zoom in for best view.}}
  }
\label{fig:supp_task_city}
\end{figure*}

We further provide qualitative evidence in \cref{fig:supp_task_city}. As demonstrated by the first example in \cref{fig:supp_task_city}, the rider and bicycle regions are detected. However, the other ``stuff" regions are misclassified in the semantic inference output when \texttt{task}=``instance". Similarly, the people are detected in the second example, and the other ``stuff" regions are misclassified. In further evidence, in both examples, the distinct ``thing" objects are segmented into a single amorphous mask in the panoptic and instance inference outputs when \texttt{task}=``semantic". Therefore, the differences in the qualitative results demonstrate \modelname's ability to output task-dependent class and mask predictions, which our task token input can guide. 

\section{Comparison to SOTA Methods at System-Level for Image Segmentation}

In this section, we compare \modelname to other SOTA systems for panoptic, instance, and semantic segmentation tasks on the ADE20K val~\cite{ade20k}, Cityscapes val~\cite{cityscapes}, and COCO val2017~\cite{coco} datasets. As shown in \cref{fig:plots}, our single \modelname model outperforms Mask2Former for the three image segmentation tasks on all three datasets.
Note that we are comparing the same OneFormer models referenced in our main text to other systems without applying additional system-level training techniques or using additional data and huge backbones.

\subsection{SOTA Systems on ADE20K val}

As shown in \cref{tab:pwc_ade20k}, without using any extra training data, Swin-L \modelname sets new state-of-the-art performance on instance segmentation with \textbf{37.8\% AP}, and DiNat-L \modelname sets new state-of-the-art performance on panoptic segmentation with \textbf{51.5\% PQ} beating the previous state-of-the-art Swin-L Mask2Former's~\cite{mask2former} 34.9\% AP and ConvNeXt-L KMaX-DeepLab's~\cite{kmax_deeplab} 50.9\% PQ, respectively. Furthermore, DiNAT-L \modelname and ConvNeXt-L \modelname achieve the new-state-of-the-art single-scale and multi-scale mIoU scores of \textbf{58.3\%} and \textbf{58.8\%}, respectively, compared to other systems that do not use extra data during training.

\subsection{SOTA Systems on Cityscapes val}

Without any extra data during training, our ConvNeXt-L \modelname sets the new state-of-the-art performance on panoptic segmentation with \textbf{68.5\% PQ} with single-scale inference. Similarly, ConvNeXt-XL \modelname achieves a new state-of-the-art \textbf{46.7\% AP} score with single-scale inference as shown in \cref{tab:pwc_cityscapes}.

\subsection{SOTA Systems on COCO val}

Without using any extra training data, DiNAT-L \modelname matches the previous state-of-the-art KMaX-DeepLab~\cite{kmax_deeplab} with \textbf{58.0\% PQ} score. Swin-L \modelname achieves the best \textbf{PQ$^\text{Th}$} score of \textbf{64.4\%}. For evaluating on the semantic segmentation task, we generate semantic GT annotations from the corresponding panoptic annotations. As shown in \cref{tab:pwc_coco}, DiNAT-L \modelname achieves an impressive \textbf{68.1\% mIoU}.

While analyzing the COCO dataset, we found serious discrepancies between the GT panoptic and instance annotations. Therefore, for fair comparison, during evaluation, we generate the instance annotations from the panoptic annotations for calculating the AP scores as only use panoptic annotations during training. We provide more information about the discrepancies in \cref{sec:coco_instance}. DiNAT-L \modelname achieves \textbf{49.2\% AP} outperforming Mask2Former-Instance~\cite{mask2former}. 

\begin{table*}[t!]\setlength{\tabcolsep}{10pt}
  \centering
  \resizebox{1.\linewidth}{!}{
  \begin{tabular}{ll|ccc|ccc|cc}
   \textbf{Method} & \textbf{Backbone} & \#\textbf{Params} & \textbf{Crop Size} & \begin{tabular}{@{}c@{}} \textbf{Extra Data} \end{tabular} & \begin{tabular}{@{}c@{}} \textbf{MS} \\ (PQ \& AP) \end{tabular} & \textbf{PQ} & \textbf{AP} & \begin{tabular}{@{}c@{}} \textbf{mIoU} \\ (s.s.) \end{tabular} & \begin{tabular}{@{}c@{}} \textbf{mIoU} \\ (m.s.) \end{tabular} \\

    \midrule
    \midrule
    \multicolumn{9}{l}{\textit{\textbf{Individual Training}}}\\
    \midrule
    \midrule

    \begin{tabular}{@{}l@{}}{HRNetV2-OCR+PSA}~\cite{polarized_hrnet} \end{tabular} & {HRNetV2-W48}~\cite{hrnet} & {---} & {$1024\!\times\!2048$} & \cmark & \begin{tabular}{@{}c@{}} {\xmark} \end{tabular} & \begin{tabular}{@{}c@{}} {---} \end{tabular} & \begin{tabular}{@{}c@{}} {---} \end{tabular} & \begin{tabular}{@{}c@{}} {---} \end{tabular} & {86.9} \\

    \begin{tabular}{@{}l@{}}{HRNetV2-OCR}~\cite{polarized_hrnet} \end{tabular} & {HRNetV2-W48}~\cite{hrnet} & {---} & {$1024\!\times\!2048$} & \cmark & \begin{tabular}{@{}c@{}} {\xmark} \end{tabular} & \begin{tabular}{@{}c@{}} {---} \end{tabular} & \begin{tabular}{@{}c@{}} {---} \end{tabular} & \begin{tabular}{@{}c@{}} {---} \end{tabular} & {86.3} \\

    \begin{tabular}{@{}l@{}}{Mask2Former}~\cite{mask2former} \end{tabular} & {ViT-Adapter-L}~\cite{chen2022vitadapter} & {571M} & {$896\!\times\!896$} & \cmark & \begin{tabular}{@{}c@{}} {\xmark} \end{tabular} & \begin{tabular}{@{}c@{}} {---} \end{tabular} & \begin{tabular}{@{}c@{}} {---} \end{tabular} & \begin{tabular}{@{}c@{}} {84.9} \end{tabular} & {85.8} \\

    \midrule
    \begin{tabular}{@{}l@{}}{Mask2Former}~\cite{mask2former} \end{tabular} & {SeMask Swin-L}~\cite{semask} & {223M} & {$512\!\times\!1024$} & \xmark & \begin{tabular}{@{}c@{}} {\xmark} \end{tabular} & \begin{tabular}{@{}c@{}} {---} \end{tabular} & \begin{tabular}{@{}c@{}} {---} \end{tabular} & \begin{tabular}{@{}c@{}} {84.0} \end{tabular} & {85.0} \\

    \begin{tabular}{@{}l@{}}Mask2Former-Semantic~\cite{mask2former} \end{tabular} & Swin-L~\cite{swin-T} & 215M & $512\!\times\!1024$ & \xmark  & \begin{tabular}{@{}c@{}} \xmark  \end{tabular} & \begin{tabular}{@{}c@{}} --- \end{tabular} & --- & \begin{tabular}{@{}c@{}} 83.3 \end{tabular} & \begin{tabular}{@{}c@{}} 84.3 \end{tabular} \\

    \midrule

    \begin{tabular}{@{}l@{}}{Panoptic-DeepLab}~\cite{panoptic-deeplab} \end{tabular} & {SWideRNet}~\cite{swidernet_2020} & {---} & {$1025\!\times\!2049$} & \cmark & \cmark & \begin{tabular}{@{}c@{}} {69.6} \end{tabular} & \begin{tabular}{@{}c@{}} {46.8} \end{tabular} & \begin{tabular}{@{}c@{}} {---} \end{tabular} & {85.3} \\

    \begin{tabular}{@{}l@{}}Axial-DeepLab-XL~\cite{axial-deeplab} \end{tabular} & Axial ResNet-XL~\cite{axial-deeplab} & 173M & $1025\!\times\!2049$ & \cmark & \cmark & \begin{tabular}{@{}c@{}} 68.5 \end{tabular} & \begin{tabular}{@{}c@{}} 44.2 \end{tabular} & \begin{tabular}{@{}c@{}} --- \end{tabular} & 84.6 \\

    \begin{tabular}{@{}l@{}}EfficientPS~\cite{mohan2020efficientps} \end{tabular} & EfficientNet~\cite{efficientnet} & --- & $1025\!\times\!2049$ & \cmark & \cmark & \begin{tabular}{@{}c@{}} 67.5 \end{tabular} & \begin{tabular}{@{}c@{}} 43.5 \end{tabular} & \begin{tabular}{@{}c@{}} --- \end{tabular} & 82.1 \\

    \midrule

     \begin{tabular}{@{}l@{}}Panoptic-DeepLab~\cite{panoptic-deeplab} \end{tabular} & SWideRNet~\cite{swidernet_2020} & --- & $1025\!\times\!2049$ & \cmark & \xmark & \begin{tabular}{@{}c@{}} 68.5 \end{tabular} & \begin{tabular}{@{}c@{}} 42.8 \end{tabular} & \begin{tabular}{@{}c@{}} 84.6 \end{tabular} & 85.3 \\

     \begin{tabular}{@{}l@{}}Axial-DeepLab-XL~\cite{axial-deeplab} \end{tabular} & Axial ResNet-XL~\cite{axial-deeplab} & 173M  & $1025\!\times\!2049$ & \cmark & \xmark & \begin{tabular}{@{}c@{}} 67.8 \end{tabular} & \begin{tabular}{@{}c@{}} 41.9 \end{tabular} & \begin{tabular}{@{}c@{}} 84.2 \end{tabular} & --- \\

    \midrule
    
    \begin{tabular}{@{}l@{}}kMaX-DeepLab~\cite{kmax_deeplab} \end{tabular} & ConvNeXt-L~\cite{convnext} & 232M & $1025\!\times\!2049$ & \xmark & \xmark & \begin{tabular}{@{}c@{}} 68.4 \end{tabular} & \begin{tabular}{@{}c@{}} 44.0 \end{tabular} & \begin{tabular}{@{}c@{}} 83.5 \end{tabular} & --- \\
    
     \begin{tabular}{@{}l@{}}Panoptic-DeepLab~\cite{panoptic-deeplab} \end{tabular} & SWideRNet~\cite{swidernet_2020} & --- & $1025\!\times\!2049$ & \xmark & \xmark & \begin{tabular}{@{}c@{}} 66.4 \end{tabular} & \begin{tabular}{@{}c@{}} 40.1 \end{tabular} & \begin{tabular}{@{}c@{}} 82.2 \end{tabular} & 82.9 \\

    \begin{tabular}{@{}l@{}}Axial-DeepLab-XL~\cite{axial-deeplab} \end{tabular} & Axial ResNet-XL~\cite{axial-deeplab} & 173M  & $1025\!\times\!2049$ & \xmark & \xmark & \begin{tabular}{@{}c@{}} 64.4 \end{tabular} & \begin{tabular}{@{}c@{}} 36.7 \end{tabular} & \begin{tabular}{@{}c@{}} 80.6 \end{tabular} & 81.1 \\
    
    \begin{tabular}{@{}l@{}}Mask2Former-Panoptic~\cite{mask2former} \end{tabular} & Swin-L~\cite{swin-T} & 216M & $512\!\times\!1024$ & \xmark & \xmark & \begin{tabular}{@{}c@{}} 66.6  \end{tabular} & \begin{tabular}{@{}c@{}} 43.6 \end{tabular} & \begin{tabular}{@{}c@{}} 82.9 \end{tabular} & \begin{tabular}{@{}c@{}} --- \end{tabular} \\
    
    \midrule
    \begin{tabular}{@{}l@{}}Mask2Former-Instance~\cite{mask2former} \end{tabular} & Swin-L~\cite{swin-T} & 216M & $512\!\times\!1024$ & \xmark  & \xmark & \begin{tabular}{@{}c@{}} ---  \end{tabular} & \begin{tabular}{@{}c@{}} 43.7 \end{tabular} & \begin{tabular}{@{}c@{}} --- \end{tabular} & \begin{tabular}{@{}c@{}} --- \end{tabular} \\
    
    \midrule
    \midrule
    \multicolumn{9}{l}{\textit{\textbf{Joint Training}}}\\
    \midrule
    \midrule
    
    \begin{tabular}{@{}l@{}}\textbf{\modelname} \end{tabular} & Swin-L~\cite{swin-T} & 219M & $512\!\times\!1024$ & \xmark & \xmark  & \begin{tabular}{@{}c@{}} \textbf{67.2} \end{tabular} & \begin{tabular}{@{}c@{}} \textbf{45.6} \end{tabular} & \begin{tabular}{@{}c@{}} 83.0 \end{tabular} & \begin{tabular}{@{}c@{}} \textbf{84.4} \end{tabular} \\
    
    \midrule
    \begin{tabular}{@{}l@{}}\textbf{\modelname} \end{tabular} & ConvNeXt-L~\cite{convnext} & 220M & $512\!\times\!1024$ & \xmark & \xmark  & \begin{tabular}{@{}c@{}} \textbf{68.5} \end{tabular} & \begin{tabular}{@{}c@{}} \textbf{46.5} \end{tabular} & \begin{tabular}{@{}c@{}} 83.0 \end{tabular} & \begin{tabular}{@{}c@{}} 84.0 \end{tabular} \\
    
    \begin{tabular}{@{}l@{}}\textbf{\modelname} \end{tabular} & ConvNeXt-XL~\cite{convnext} & 372M & $512\!\times\!1024$ & \xmark & \xmark  & \begin{tabular}{@{}c@{}} \textbf{68.4} \end{tabular} & \begin{tabular}{@{}c@{}} \textbf{46.7} \end{tabular} & \begin{tabular}{@{}c@{}} \textbf{83.6} \end{tabular} & \begin{tabular}{@{}c@{}} \textbf{84.6} \end{tabular} \\
    
    \midrule
    \begin{tabular}{@{}l@{}}\textbf{\modelname} \end{tabular} & DiNAT-L~\cite{dinat} & 223M & $512\!\times\!1024$ & \xmark & \xmark  & \begin{tabular}{@{}c@{}} \textbf{67.6} \end{tabular} & \begin{tabular}{@{}c@{}} \textbf{45.6} \end{tabular} & \begin{tabular}{@{}c@{}} 83.1 \end{tabular} & \begin{tabular}{@{}c@{}} 84.0 \end{tabular} \\
    
    \bottomrule
    
\end{tabular}}
  \vspaceundertab
  \caption{\textbf{Comparison to SOTA systems on Cityscapes val~\cite{cityscapes}.} \modelname achieves new-state-of-the-art performances on the instance and panoptic segmentation tasks when compared with SOTA systems \textbf{using single-scale inference}.
      }
    \label{tab:pwc_cityscapes}
\end{table*}

\begin{table*}[t!]\setlength{\tabcolsep}{10pt}
  \centering
  \resizebox{1.\linewidth}{!}{
  \begin{tabular}{ll|cc|ccc|cc|c}
  \textbf{Method} & \textbf{Backbone} & \#\textbf{Params} & \textbf{Extra Data} & \textbf{PQ} & \textbf{PQ}$^\text{Th}$ & \textbf{PQ}$^\text{St}$ & \textbf{AP} & \gr{\textbf{AP}$^\text{instance}$} & \begin{tabular}{@{}c@{}} \textbf{mIoU} \end{tabular} \\

   \midrule
    \midrule
    \multicolumn{9}{l}{\textit{\textbf{Individual Training}}}\\
    \midrule
    \midrule

    \begin{tabular}{@{}l@{}}Mask DINO~\cite{li2022mask-dino} \end{tabular} & Swin-L~\cite{swin-T} & 223M & \cmark & \begin{tabular}{@{}c@{}} 59.4 \end{tabular} & \begin{tabular}{@{}c@{}} --- \end{tabular} & --- & --- & \begin{tabular}{@{}c@{}} \gr{54.5} \end{tabular} & --- \\

    \begin{tabular}{@{}l@{}}kMaX-DeepLab~\cite{kmax_deeplab} \end{tabular} & ConvNeXt-L~\cite{convnext} & 232M & \cmark & 58.1 & \begin{tabular}{@{}c@{}} 64.3 \end{tabular} & \begin{tabular}{@{}c@{}} 48.8\end{tabular} & --- & \gr{---} & --- \\

    \midrule
    \begin{tabular}{@{}l@{}}kMaX-DeepLab~\cite{kmax_deeplab} \end{tabular} & ConvNeXt-L~\cite{convnext} & 232M & \xmark & 58.0 & \begin{tabular}{@{}c@{}} 64.2 \end{tabular} & \begin{tabular}{@{}c@{}} 48.6\end{tabular} & --- & \gr{---} & --- \\

    \begin{tabular}{@{}l@{}}Mask2Former-Panoptic~\cite{mask2former} \end{tabular} & Swin-L~\cite{swin-T} & 216M & \xmark  & \begin{tabular}{@{}c@{}} 57.8 \end{tabular} & \begin{tabular}{@{}c@{}} 64.2  \end{tabular} & \begin{tabular}{@{}c@{}} 48.1
    \end{tabular} & \begin{tabular}{@{}c@{}} 48.7 \end{tabular} & \gr{48.6} & \begin{tabular}{@{}c@{}} 67.4 \end{tabular} \\
    
    \begin{tabular}{@{}l@{}} Panoptic SegFormer~\cite{panoptic-segformer} \end{tabular} & Swin-L~\cite{swin-T} & 221M & \xmark  &  55.8 & \begin{tabular}{@{}c@{}} 61.7 \end{tabular} & \begin{tabular}{@{}c@{}} 46.9 \end{tabular} & --- & \gr{---} & --- \\

    \midrule   
    
    \begin{tabular}{@{}l@{}}Mask2Former-Instance~\cite{mask2former} \end{tabular} & Swin-L~\cite{swin-T} & 216M & \xmark  & \begin{tabular}{@{}c@{}} --- \end{tabular} & \begin{tabular}{@{}c@{}} ---  \end{tabular} & \begin{tabular}{@{}c@{}} ---
    \end{tabular} & \begin{tabular}{@{}c@{}} \textbf{49.1} \end{tabular} & \gr{50.1} & \begin{tabular}{@{}c@{}} --- \end{tabular} \\
    
    \midrule
    \midrule
    \multicolumn{9}{l}{\textit{\textbf{Joint Training}}}\\
    \midrule
    \midrule
    
    \begin{tabular}{@{}l@{}}\textbf{\modelname} \end{tabular} & Swin-L~\cite{swin-T} & 219M & \xmark  & \begin{tabular}{@{}c@{}} \textbf{57.9} \end{tabular} & \begin{tabular}{@{}c@{}} \textbf{64.4} \end{tabular} & \begin{tabular}{@{}c@{}} 48.0  \end{tabular} & \begin{tabular}{@{}c@{}} \textbf{49.0} \end{tabular} & \gr{48.9} & \begin{tabular}{@{}c@{}} \textbf{67.4} \end{tabular} \\
    \midrule
    
    \begin{tabular}{@{}l@{}}\textbf{\modelname} \end{tabular} & DiNAT-L~\cite{dinat} & 223M & \xmark  & \begin{tabular}{@{}c@{}} \textbf{58.0} \end{tabular} & \begin{tabular}{@{}c@{}} \textbf{64.3} \end{tabular} & \begin{tabular}{@{}c@{}} \textbf{48.4}  \end{tabular} & \begin{tabular}{@{}c@{}} \textbf{49.2} \end{tabular} & \gr{49.2} & \begin{tabular}{@{}c@{}} \textbf{68.1} \end{tabular} \\
    
    \bottomrule
\end{tabular}}
  \vspaceundertab
  \caption{\textbf{Comparison to SOTA systems on COCO val2017~\cite{coco}.} OneFormer achieves the best PQ$^\text{Th}$ score among the SOTA systems trained without using any extra data. \gr{AP$^\text{instance}$ represents evaluation on the original instance annotations.} 
      }
    \label{tab:pwc_coco}
    \vspaceundercaption
\end{table*}

\section{Analysis on Discrepancy between Instance and Panoptic Annotations in COCO}
\label{sec:coco_instance}

\begin{figure*}[t!]
\centering
\includegraphics[width=1.0\linewidth]{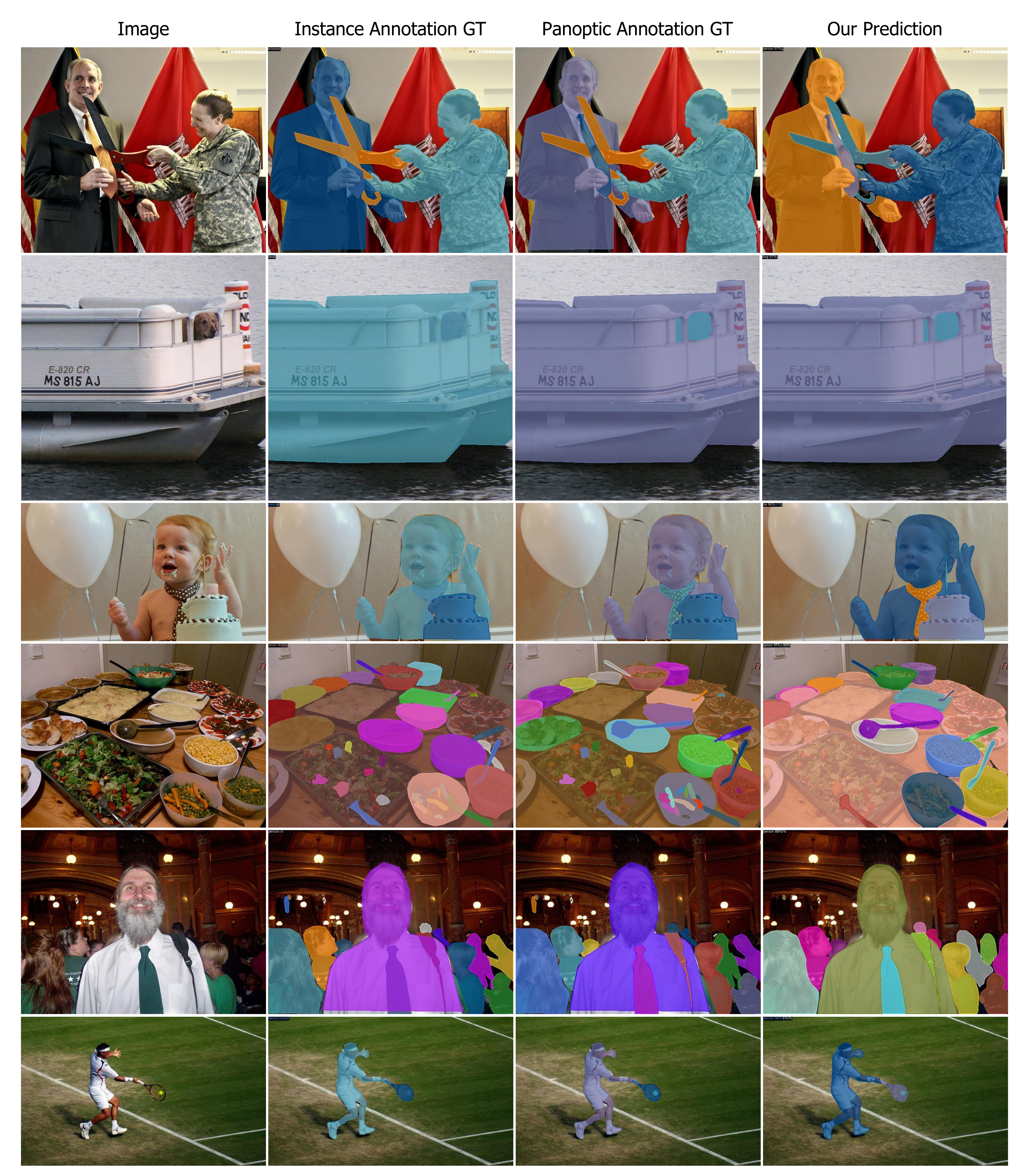}
\caption{
  \textbf{Discrepancy between instance and panoptic annotations in the COCO train2017~\cite{coco} dataset.} The ``tie" instance is merged into the ``person" instance in the instance annotations, whereas the panoptic annotations segment the two objects separately in the first, third, and fifth rows. Similarly, ``dog" and ``boat" are merged into a single instance in the instance annotations in the second row. The ``bowl" and ``spoon" are segmented as a single instance in instance annotations in the fourth row. Lastly, the `tennis racket" and the small ``sports ball" are segmented distinctly in panoptic annotations, unlike instance annotations in the last row. \textbf{\pp{Zoom in for best view.}}
  }
\label{fig:supp_coco_train}
\end{figure*}

\begin{figure*}[t!]
\centering
\includegraphics[width=1.0\linewidth]{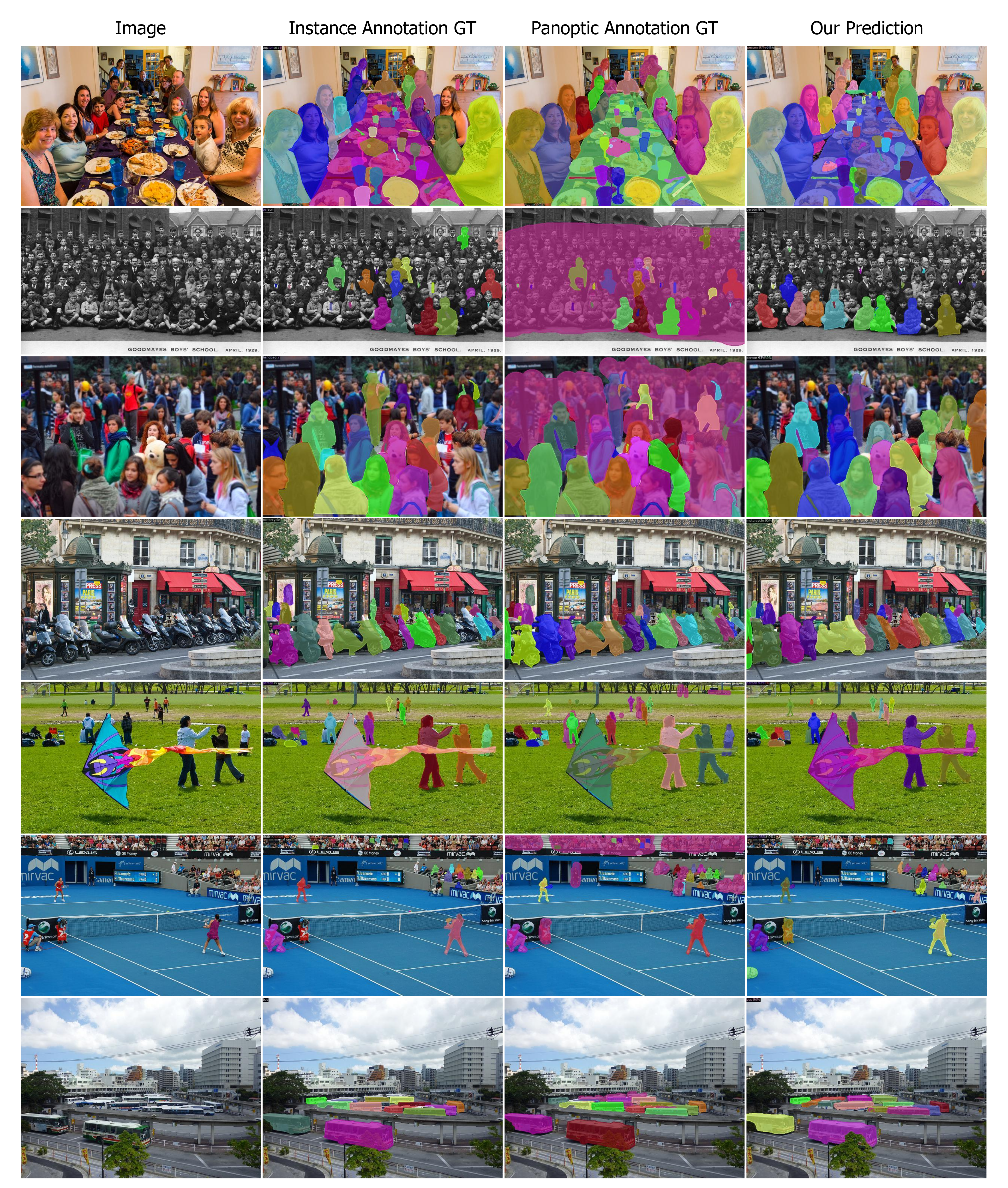}
\caption{
  \textbf{Discrepancy between instance and panoptic annotations in the COCO val2017~\cite{coco} dataset.} The instance annotations skip multiple ``person" and ``motorcycle" objects in the first and fourth rows. The instance annotations leave out a group of ``person" objects in the background, and panoptic annotations merge those objects into a single object mask in the second, third, fifth, and sixth rows. A similar case is observed with ``bus" in the background in the last row.  \textbf{\pp{Zoom in for best view.}}}
\label{fig:supp_coco_val}
\end{figure*}

During our joint training, we derive the semantic and instance ground-truth labels from the corresponding panoptic annotations. Unlike, Cityscapes~\cite{cityscapes} and ADE20K~\cite{ade20k} datasets, which combine the semantic and instance annotations to generate the corresponding panoptic annotations while preparing the data, COCO~\cite{coco} has separate sets of panoptic and instance annotations. As expected, there are no discrepancies between the panoptic and instance annotations in the Cityscapes~\cite{cityscapes} and ADE20K~\cite{ade20k} datasets. However, because COCO~\cite{coco} has separately developed panoptic and instance annotations, we discover significant discrepancies in the COCO train2017 and val2017~\cite{coco} datasets as shown in \cref{fig:supp_coco_train} and \cref{fig:supp_coco_val}, respectively.

In \cref{fig:supp_coco_train}, the instance annotations merge the ``tie" object into the ``person" object. In another example, instance annotations merge the ``dog" and ``boat" into a single instance, while the panoptic annotations segment the two instances correctly.

In \cref{fig:supp_coco_val}, the instance annotations skip multiple ``person" and ``motorcycle" objects in different images, while the panoptic annotations include them all. In another example, instance annotations leave out a group of ``person" object instances in the background, and panoptic annotations merge those instances into a single object mask.

These discrepancies are a significant barrier to developing and evaluating a unified image segmentation model. As demonstrated in \cref{fig:supp_coco_train} and \cref{fig:supp_coco_val}, our predictions match the panoptic annotations much more than the instance annotations which is expected from our training strategy involving only panoptic annotations. Therefore, while comparing our Swin-L$^\dag$ \modelname to other SOTA methods in Tab. \pp{3} (main text), we evaluate the AP score on instance GTs derived from the panoptic annotations.

\end{document}